\documentclass[11pt]{article}
\newif\ifanonymous
\ifanonymous\usepackage[review]{acl}\else\usepackage[final]{acl}\fi
\usepackage{times}
\usepackage{latexsym}
\usepackage[T1]{fontenc}
\usepackage[utf8]{inputenc}
\usepackage{microtype}
\usepackage{inconsolata}
\usepackage{booktabs,tabularx,array,graphicx}
\usepackage{multirow}
\usepackage{pifont}
\usepackage{colortbl}
\usepackage{amsmath,amssymb}
\usepackage{fvextra}
\usepackage{enumitem}
\ifdefined\XeTeXversion
\usepackage{newunicodechar}
\newunicodechar{ă}{\u{a}}
\newunicodechar{Ł}{\L{}}
\newunicodechar{ł}{\l{}}
\newunicodechar{’}{'}
\else
\DeclareUnicodeCharacter{0103}{\u{a}}
\DeclareUnicodeCharacter{0141}{\L{}}
\DeclareUnicodeCharacter{0142}{\l{}}
\DeclareUnicodeCharacter{2019}{'}
\fi
\definecolor{BestBg}{HTML}{E2EFDA}
\definecolor{SecondBg}{HTML}{DDEBF7}
\newcommand{\cmark}{\ding{51}}
\newcommand{\xmark}{\ding{55}}
\usepackage{xspace}
\newcommand{\ours}{\textsc{InfoOrgBench}\xspace}

\newcommand{\github}{\raisebox{-1.5pt}{\includegraphics[height=1.05em]{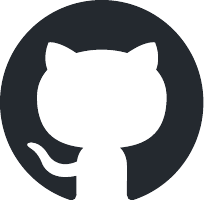}}\xspace}
\newcommand{\startbackmatter}{%
  \par
  \ifnum\value{page}>9
    \PackageError{InfoOrgBench}{Main content exceeds eight physical pages}{Move secondary material to the appendix before compiling.}%
  \fi
}

\title{InfoOrgBench: Can Agents Turn Scattered Information into Coherent Hierarchies?}
\ifanonymous

\newcommand{\codelink}{\href{https://anonymous.4open.science/r/sub-InforOrgBench-3C08/}{\path{sub-InforOrgBench-3C08}}}
\author{Anonymous authors}
\else

\newcommand{\codelink}{\href{https://github.com/microsoft/InfoOrgBench}{\path{microsoft/InfoOrgBench}}}
\author{
  Justice Ou\textsuperscript{1,2}\begin{NoHyper}\thanks{Corresponding author: \href{mailto:zo6@illinois.edu}{zo6@illinois.edu}.}\end{NoHyper},
  Lanming Ma\textsuperscript{3},
  Zijian Zhang\textsuperscript{4},
  Yilun Zhao\textsuperscript{5}
\\
  \textsuperscript{1}Microsoft,
  \textsuperscript{2}University of Illinois Urbana-Champaign
\\
  \textsuperscript{3}New York University,
  \textsuperscript{4}University of Michigan,
  \textsuperscript{5}Yale University
}
\fi
\begin{document}
\maketitle

\begin{abstract}
When agents organize records into folders, they may correctly group the excerpts of a design proposal yet place the proposal under the wrong owning group. Existing agent benchmarks do not directly measure such higher-level errors.
We introduce \ours, a benchmark of 150 tasks in which agents organize scattered records from meeting recaps, issue threads and design proposals into hierarchies. Integrity checks verify that every record is preserved, and a Structure Score measures agreement with the source hierarchy at every level, independent of folder names and order.
We evaluate 45 systems that pair nine models with five agent harnesses. The best system scores 83.2 out of 100, but no system exceeds 78 on meeting recaps or issue threads. Model choice affects scores far more than harness choice: model means span about 27 points, whereas the average effect of each harness is below one point. More importantly, parent-level relations have higher error rates than leaf-level relations in all three domains. In design proposals, nearly half of the outputs analyzed by level make exactly this kind of error. These results show that evaluating information organization requires measuring higher-level relations separately from record preservation and local grouping.

\begin{small}
\begin{center}
\begin{tabular}{cll}
\github & \textbf{Code} & \codelink\\
\end{tabular}
\end{center}
\end{small}
\end{abstract}

\section{Introduction}
\label{sec:introduction}
\begin{figure*}[t]
\centering
\includegraphics[width=\textwidth]{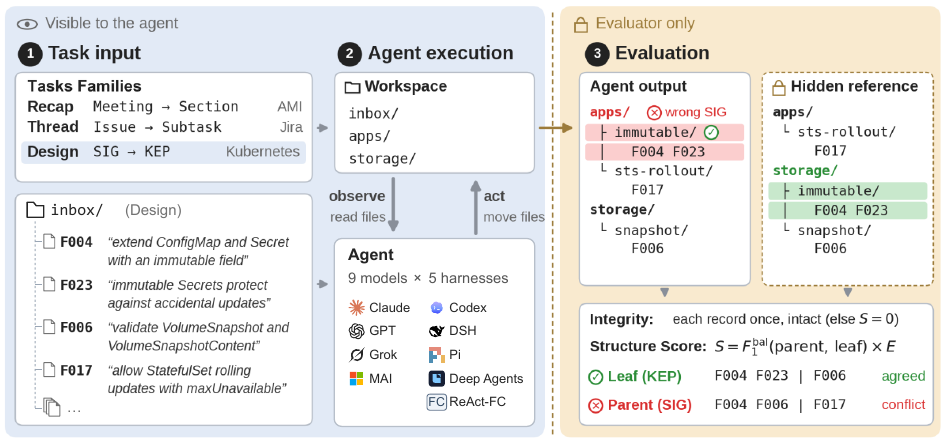}
\caption{\textbf{Overview of \ours.} Agents see records with opaque IDs and an organizing criterion, and reorganize them through harness tools; the final files are captured when the agent stops. In this Design example, the agent keeps the immutable-secrets KEP intact but files it under the wrong SIG: the leaf-level triple agrees with the hidden reference, while the parent-level triple conflicts. Valid outputs are scored by the Structure Score $S$, which ignores folder names; $E$ penalizes excess single-child folders. KEPs are Kubernetes Enhancement Proposals, grouped by their owning Special Interest Groups (SIGs).}\label{fig:F1}
\end{figure*}

Agents increasingly collect and produce records for people, such as meeting notes, issue discussions and design documents. Keeping these records useful requires more than putting related items together: each group must also sit under the right parent, because people rely on folder hierarchies both to find information later and to plan how a project divides into parts \citep{ref-folders}. The two requirements can come apart. An agent may gather every excerpt of a design proposal into one folder yet file the proposal under the wrong owning group. Such an output keeps every record, so an inventory check would count it as complete, whereas a binary exact-match criterion would score it as the same failure as an output whose entire hierarchy is disrupted. Evaluating organization therefore requires separate, graded measures of record preservation, local grouping and the higher-level relations that connect groups.

Existing agent benchmarks mainly evaluate task completion, such as resolving repository issues, operating websites, desktops and enterprise software, or working with large file collections \citep{ref-swebench,ref-webarena,ref-osworld,ref-workarena,ref-workspace}. Work on information organization uses language models to cluster texts, build taxonomies and route documents into an existing folder hierarchy \citep{zhang-etal-2023-clusterllm,ref-tnt-llm,ref-paperrouter}. These settings do not directly isolate whether an agent, given a fixed collection of records, recovers both the local groups and the parent relations that connect them. We therefore ask how reliably agents recover the relations between groups of records, beyond preserving the records or clustering them.

We introduce \ours, a benchmark of 150 tasks in which agents organize scattered records into folders. The tasks span three families built from public sources (Figure~\ref{fig:F1}): meeting recaps (Recap) group AMI summary sentences \citep{ref-ami} by meeting and section, issue threads (Thread) group Jira records by parent issue and subtask, and design proposals (Design) group Kubernetes proposal excerpts \citep{ref-kubernetes} by owning group and proposal. Agents see the record contents and an organizing criterion, while the reference hierarchy recorded in the sources remains with the evaluator. After each episode, final-state checks confirm that every record is kept unchanged and filed, and a Structure Score compares the output with the reference at every level of the hierarchy, independent of folder names and order, while penalizing excess single-child folders. Scores therefore measure agreement with the source hierarchy, not whether other organizations of the same records would also be useful.

Each agent works through a harness, which sets its tools, the history it keeps and when the episode ends, so we evaluate all 45 systems that pair nine models with five agent harnesses. GPT-6.1 Sol achieves the highest score under every harness, reaching 83.2 out of 100; by family, no system exceeds 78 on Recap or Thread, whereas the best reaches 95.8 on Design. The choice of model affects scores far more than the choice of harness: model means span about 27 points, whereas the average effect of each harness is below one point. Analyzing errors by hierarchy level, we find that parent-level relations have higher error rates than leaf-level relations in every family. In Recap, more than half of the output folders mix notes from different meetings; in Thread, half of the outputs omit the parent-issue layer; in Design, nearly half of the outputs analyzed by level keep every proposal group intact but place some under the wrong owning group. These results indicate that the evaluated agents recover local groups more reliably than the hierarchy that connects them.

We make the following contributions:
\begin{itemize}[leftmargin=*]
  \item We introduce \ours, a benchmark of 150 tasks in which agents organize scattered records into hierarchies, with references constructed deterministically from hierarchies recorded in public sources.
  \item We design an evaluation that checks record preservation separately from structure and scores structural agreement at each hierarchy level with a Structure Score that ignores folder names and penalizes excess single-child folders.
  \item We evaluate 45 systems that pair nine models with five harnesses and find that, in every task family, agents recover parent-level relations less reliably than leaf-level groups, and that model choice affects scores far more than harness choice.
\end{itemize}

\section{Related Work}
\label{sec:related}
\begin{table*}[t]
\centering
\small
\setlength{\tabcolsep}{4pt}
{\scriptsize
\setlength{\tabcolsep}{1.5pt}
\begin{tabular*}{\linewidth}{@{\extracolsep{\fill}}lccc*{5}{c}l@{}}
\toprule
\textbf{Benchmark} & \textbf{Tasks} & \textbf{Input per task} & \textbf{Agent actions} & \textbf{Files} & \textbf{No VM} & \textbf{Organizes} & \textbf{Prior ref.} & \textbf{LLM-free eval.} & \textbf{Reference} \\
\midrule
\multicolumn{10}{@{}l}{\textit{Agent benchmarks}}\\
WebArena & 812 & Websites & Click, type & \xmark & \xmark & \xmark & \xmark & \xmark & Author checks \\
OSWorld & 369 & OS VM & Mouse, keys & \xmark & \xmark & \xmark & \xmark & \cmark & Eval scripts \\
SWE-bench & 2,294 & \textasciitilde{}3k-file repo & Write patch & \cmark & \xmark & \xmark & \cmark & \cmark & PR tests \\
Workspace-Bench & 388 & \textasciitilde{}4k files & Shell, files & \cmark & \xmark & \xmark & \xmark & \xmark & Rubrics \\
\midrule
\multicolumn{10}{@{}l}{\textit{Hierarchy benchmarks}}\\
HSCodeComp & 632 & Product & Search & \xmark & \cmark & \xmark & \xmark & \cmark & Expert codes \\
TExEval-2 & 24 & Term list & None & \xmark & \cmark & \cmark & \cmark & \cmark & Gold taxonomies \\
\midrule
\multicolumn{10}{@{}l}{\textit{This work}}\\
\textbf{\ours} & \textbf{150} & \textbf{40 \textasciitilde{} 60 files} & \textbf{Read, search, move} & \cmark & \cmark & \cmark & \cmark & \cmark & \textbf{Source hierarchy} \\
\bottomrule
\end{tabular*}
}\par\vspace{3pt}
\caption{Comparison with existing benchmarks. \ours asks an agent to move every inbox record into folders that reconstruct its source hierarchy, such as meeting / section / sentence. Files indicates filesystem interaction; No VM indicates that no virtual machine, container or hosted service is required; Organizes indicates that the output groups multiple supplied items; Prior ref. indicates that the reference predates the benchmark; LLM-free eval. indicates that scoring invokes no language model. \ours scores source-hierarchy relations and ignores folder names.}\label{tab:T1}
\end{table*}

\paragraph{Agent and Workspace Benchmarks.} Benchmarks evaluate code generation, completion and repository issue resolution \citep{ref-humaneval,ref-mbpp,ref-apps,ref-repobench,ref-crosscodeeval,ref-swebench}, agents operating websites, enterprise software and operating systems \citep{ref-webarena,ref-workarena,ref-workarena-plus,ref-agentcompany,ref-osworld,wei-etal-2026-opencomputer,ref-agentbench}, and tasks over interdependent files in large workspaces \citep{ref-workspace}; results also depend on the harness and execution interface as well as the model \citep{ref-harness-bench,zhou-etal-2026-gui}. These benchmarks score task completion rather than the resulting folder hierarchy.

\paragraph{Clustering and Hierarchy Induction.} Language models cluster texts and build label taxonomies \citep{zhang-etal-2023-clusterllm,ref-tnt-llm}; \citet{zhu-etal-2025-context} compare generated paper taxonomies with those from surveys, using flat clustering metrics and heading similarity. Related work extracts single-document hierarchies \citep{ref-held} or compares term taxonomies with gold ones \citep{ref-texeval}. These evaluations treat groupings as flat partitions or compare structures through labels; \ours scores record relations at each hierarchy level, independent of folder names.

\paragraph{Applying and Using Hierarchies.} PaperRouter-Agent routes papers into a user's existing folder hierarchy \citep{ref-paperrouter}, and HSCodeComp asks agents to assign products to hierarchical tariff codes \citep{ref-hscodecomp}. Other work studies how organized information supports exploration \citep{ref-sensecape}, agent memory \citep{ref-fsmemory}, agentic search \citep{zhao-etal-2026-rethinking}, question answering over documents and patient records \citep{ref-madqa,ou-etal-2026-experience,shen-etal-2026-patient}, and research synthesis \citep{ref-storm,ref-deepresearch,zhao-etal-2025-sciarena}. These settings evaluate placement within a given hierarchy or downstream use; \ours evaluates the constructed hierarchy itself against the source.

In Table~\ref{tab:T1}, only \ours has agents organize supplied items through file operations and scores them against a pre-existing reference.

\section{\texorpdfstring{\ours}{InfoOrgBench} Benchmark}
\label{sec:benchmark}

This section defines the organization task, describes how tasks and their reference hierarchies are built from public sources, and characterizes the resulting data and the dependence among tasks.
\begin{table*}[htbp]
\centering
\small
\setlength{\tabcolsep}{4pt}
\begin{tabularx}{\textwidth}{@{}Xrrrrrrr@{}}
\toprule
\textbf{Family} & \textbf{Tasks} & \textbf{Records} & \textbf{Per task} & \textbf{Top groups} & \textbf{Lower groups} & \textbf{Unique top} & \textbf{Unique lower} \\ \midrule
Recap & 50 & 2,436 & 40 to 60 & 2 to 4 & 6 to 15 & 124 & 450 \\
Thread & 50 & 2,421 & 41 to 60 & 2 to 3 & 3 to 17 & 3 & 42 \\
Design & 50 & 1,800 & 36 to 36 & 4 to 4 & 12 to 12 & 16 & 294 \\
\bottomrule
\end{tabularx}
\caption{Declared scored-input inventory verified against retained packet/reference file hashes: ReAct-FC Recap and Thread, and the shared Design bundle. Top/lower groups are meetings/sections, parent issues/subtasks and owners/proposals. Ranges are per packet; unique counts use full source paths. All Design packets have four owners, three proposals per owner and three excerpts per proposal. The distinct 7,331-appearance materialization is in Table~\ref{tab:appendix-materialized-inventory}. These counts do not establish source independence or executed access equivalence.}
\label{tab:T2}
\end{table*}

\subsection{Task}
\label{sec:benchmark.task}
 
The task maps a flat or scattered collection into a hierarchy. Inputs provide stable record identifiers, an organizing criterion and task-specific evidence. Leaves retain every input record exactly once; internal nodes express groups and subgroups. An agent inspects and moves records through harness tools, and independent capture makes the final workspace the scored artifact. Structural comparison ignores consistent label renaming and sibling order. Appendix~\ref{sec:appendix.tasks} gives the evidence boundaries and worked task examples.

Design groups substantive proposal excerpts by owning group and proposal. Recap groups summary sentences by source meeting and section; Thread groups issue and comment records by parent issue and subtask. Source metadata fixes the hidden target, while supplied text and the criterion define the intended task evidence. Generated prose lies outside the output contract.

\subsection{Construction}
\label{sec:benchmark.construction}
 
Construction selects source records while retaining source-native relations. Kubernetes Enhancement Proposals provide documents, ownership metadata and heading paths. AMI links summary sentences to meetings and sections; Jira links comments and subtasks to parent issues. Extraction fixes these paths before assigning opaque record identifiers. The scored development suite has no established unique-reference admission filter or completed semantic parent-validation stage.
Appendix~\ref{sec:appendix.materialization} specifies the source versions, record extraction and task selection. Rebuilding one Recap task and one Design task from the pinned sources reproduces their packet and reference files exactly.

\subsection{Data}
\label{sec:benchmark.data}

Recap uses AMI meeting summaries. Table~\ref{tab:T2} describes retained declared inputs with 2,436 Recap, 2,421 Thread and 1,800 Design record appearances. Every Design packet has 36 records, four owners and 12 proposals: 600 proposal appearances across 294 unique proposals. Recap covers 124 meetings; Thread reuses three parent issues.

A separate 7,331-appearance materialization is kept apart from these counts.

The public sources retain their licenses. Source manifests record provenance and exclusions following the documentation principles of Datasheets for Datasets \citep{ref-datasheets}; the task appendix also states the artifact-specific license, attribution and redistribution boundaries.

\subsection{Reference Construction}
\label{sec:benchmark.validation}
 
The evaluator reference is reconstructed deterministically from the pinned source hierarchy: meeting and summary-section membership for Recap, parent issue and subtask membership for Thread, and owning group and proposal paths for Design. Materialization assigns opaque record identifiers only after these relations are fixed and stores source-native identifiers, paths and content hashes outside the agent workspace. Replay checks verify the packet and reference bytes before scoring. The resulting gold target measures reconstruction of the source organization; alternative useful organizations remain a separate question of downstream utility \citep{ref-construct-validity}.

\subsection{Dependence}
\label{sec:benchmark.splits}
 
The scored packets are overlapping development instances, without a held-out source-disjoint split. A packet can contain several recurring source units, so record appearances and task counts are not independent sample sizes. Source identifiers support dependence-aware descriptions.

\section{Evaluation}
\label{sec:evaluation}

This section defines how outputs are scored and describes the harnesses, the systems and the protocol under which they are run.
 
\subsection{Metrics}
\label{sec:evaluation.metrics}

We compare each output hierarchy $O$ with the reference hierarchy $R$ over the same records. An output is valid if every record appears exactly once with unchanged content. Invalid outputs score 0.

Agreement is measured with rooted triples, which compare hierarchies through record identities and ignore folder names \citep{ref-tree-comparison,ref-triples-distance}. A hierarchy resolves records $a$, $b$ and $c$ as $ab|c$ when $a$ and $b$ share a deeper common ancestor than either shares with $c$. Below a shared wrapper folder, every reference has two levels: parent groups (meetings, parent issues or SIGs) and leaf groups within a parent. Each triple belongs to the level at which its records separate in $R$. At each level, recall is the fraction of triples resolved by $R$ that $O$ resolves the same way, and precision is the fraction of triples resolved by $O$ that agree with $R$. Level-balanced F1, $F_1^{\mathrm{bal}}$, is the harmonic mean of recall and precision, each averaged over levels. Averaging over levels keeps the more numerous parent-level triples from dominating the score \citep{ref-hierarchical-clustering,ref-level-weighting,ref-texeval,ref-macroaveraging}. A flat output resolves no triple, so $F_1^{\mathrm{bal}}=0$. Appendix~\ref{sec:appendix.metrics} gives formal definitions and edge cases.

Folder economy $E$ penalizes single-child folders in excess of the reference \citep{ref-folder-navigation,ref-navigation-depth}:
\begin{equation}
E(O,R) = 1 - \lambda \frac{\max\{0,\,u(O)-u(R)\}}{n(O)},
\label{eq:folder-economy}
\end{equation}
where $u(\cdot)$ counts single-child folders, $n(O)$ counts the folders in $O$ and $\lambda=0.2$. The primary metric, Structure Score, combines the two:
\begin{equation}
S(O,R) = F_1^{\mathrm{bal}}(O,R)\,E(O,R).
\label{eq:structure-score}
\end{equation}
We also report pooled Triplet F1 over all triples and the economy penalty $100(1-E)$. All scores are scaled to 0--100.

\subsection{Harness}
\label{sec:evaluation.harness}
Agent performance depends on both the model and the harness, the layer around the model that manages tools, context and episode termination \citep{ref-harness-bench}. We therefore evaluate each model with five harnesses and score the final workspace state \citep{ref-osworld}.

We implement ReAct-FC, a ReAct agent \citep{ref-react} that acts through function calls. It provides five tools for listing, reading, searching and moving records and for checking the inventory. The model keeps the full episode history, including its reasoning items.

We also evaluate four public harnesses: OpenAI's Codex CLI \citep{ref-codex-cli}, DeepSeek Harness (DSH) \citep{ref-deepseek-harness}, Pi \citep{ref-pi} and Deep Agents \citep{ref-deepagents}. Each uses its own tools, built-in instructions and history management. In every harness, the final file paths and contents are captured outside the agent and define the outcome. Appendices~\ref{sec:appendix.prompts} and~\ref{sec:appendix.settings} give the task instructions and run settings.

\subsection{Systems}
\label{sec:evaluation.systems}
\begin{table*}[t]
\centering
{\scriptsize
\setlength{\tabcolsep}{1.2pt}
\renewcommand{\arraystretch}{0.75}
\begin{tabular*}{\textwidth}{@{\extracolsep{\fill}}>{\raggedright\arraybackslash}p{0.13\textwidth}>{\raggedright\arraybackslash}p{0.1\textwidth}*{12}{>{\raggedleft\arraybackslash}p{0.046\textwidth}}@{}}
\toprule
\multirow{2}{*}{\textbf{Model}} & \multirow{2}{*}{\textbf{Harness}} & \multicolumn{3}{c}{\textbf{Recap}} & \multicolumn{3}{c}{\textbf{Thread}} & \multicolumn{3}{c}{\textbf{Design}} & \multicolumn{3}{c}{\textbf{Avg}} \\
\cmidrule(lr){3-5}\cmidrule(lr){6-8}\cmidrule(lr){9-11}\cmidrule(lr){12-14}
 &  & \textbf{Trip.} & \textbf{Pen.}$\downarrow$ & \textbf{Score} & \textbf{Trip.} & \textbf{Pen.}$\downarrow$ & \textbf{Score} & \textbf{Trip.} & \textbf{Pen.}$\downarrow$ & \textbf{Score} & \textbf{Trip.} & \textbf{Pen.}$\downarrow$ & \textbf{Score} \\
\midrule
\multirow{5}{0.13\textwidth}{\raggedright Claude Haiku 4.5} & \mbox{ReAct-FC} & 38.5 & 0.8 & 40.8 & 42.1 & 9.9 & 46.2 & 76.0 & 2.9 & 79.9 & 52.2 & 4.5 & 55.6 \\
 & \mbox{Pi} & 37.3 & 1.1 & 39.8 & 50.6 & 7.0 & 53.0 & 78.4 & 2.6 & 80.9 & 55.4 & 3.6 & 57.9 \\
 & \mbox{Deep Agents} & 37.5 & 1.0 & 40.2 & 47.1 & 8.2 & 50.7 & 75.5 & 3.0 & 79.0 & 53.4 & 4.1 & 56.7 \\
 & \mbox{DSH} & 33.9 & 0.9 & 37.1 & 37.4 & 14.7 & 46.8 & 71.4 & 7.5 & 74.7 & 47.6 & 7.7 & 52.9 \\
 & \mbox{Codex} & 35.1 & 0.9 & 38.7 & 41.8 & 8.8 & 43.9 & 71.1 & 3.7 & 74.9 & 49.3 & 4.5 & 52.5 \\
\midrule
\multirow{5}{0.13\textwidth}{\raggedright Claude Haiku 5.5} & \mbox{ReAct-FC} & 39.0 & 1.6 & 39.9 & 64.3 & 3.8 & 65.1 & 88.7 & 1.0 & 90.0 & 64.0 & 2.1 & 65.0 \\
 & \mbox{Pi} & 35.9 & 1.3 & 36.9 & 66.5 & 1.5 & 66.8 & 90.8 & 0.7 & 92.0 & 64.4 & 1.2 & 65.3 \\
 & \mbox{Deep Agents} & 37.5 & 0.9 & 40.2 & 68.7 & 2.1 & 67.6 & 91.1 & 0.9 & 91.9 & 65.8 & 1.3 & 66.5 \\
 & \mbox{DSH} & 36.4 & \cellcolor{BestBg}\textbf{0.5} & 42.8 & 62.8 & 2.7 & 63.6 & 90.1 & 0.9 & 91.2 & 63.1 & 1.3 & 65.9 \\
 & \mbox{Codex} & 36.9 & 1.3 & 40.1 & 65.2 & 2.1 & 64.8 & 93.6 & 0.6 & 94.2 & 65.2 & 1.3 & 66.3 \\
\midrule
\multirow{5}{0.13\textwidth}{\raggedright Claude Sonnet 5.5} & \mbox{ReAct-FC} & 50.5 & 1.1 & 48.5 & 73.2 & 0.9 & 73.5 & 91.4 & 0.6 & 92.9 & 71.7 & 0.9 & 71.6 \\
 & \mbox{Pi} & 50.9 & 1.1 & 48.9 & 72.4 & 0.7 & 73.6 & 91.3 & 0.6 & 92.7 & 71.5 & 0.8 & 71.7 \\
 & \mbox{Deep Agents} & 51.9 & 0.8 & 50.3 & 73.6 & \cellcolor{SecondBg}\underline{0.5} & 73.5 & 92.2 & 0.8 & 93.3 & 72.6 & 0.7 & 72.4 \\
 & \mbox{DSH} & 49.6 & 1.0 & 48.9 & 72.1 & 0.9 & 72.4 & 90.8 & 0.8 & 92.0 & 70.8 & 0.9 & 71.1 \\
 & \mbox{Codex} & 48.4 & 0.9 & 47.5 & 70.9 & 1.2 & 71.5 & 90.9 & 0.7 & 92.6 & 70.0 & 0.9 & 70.5 \\
\midrule
\multirow{5}{0.13\textwidth}{\raggedright MAI-Code-1.1-Flash} & \mbox{ReAct-FC} & 36.7 & 1.1 & 41.7 & 47.3 & 5.2 & 47.2 & 77.7 & 3.2 & 79.3 & 53.9 & 3.2 & 56.1 \\
 & \mbox{Pi} & 35.8 & 1.4 & 39.8 & 54.7 & 6.1 & 54.0 & 77.1 & 3.8 & 78.0 & 55.9 & 3.8 & 57.3 \\
 & \mbox{Deep Agents} & 33.9 & 1.1 & 37.6 & 53.5 & 6.3 & 53.1 & 76.8 & 3.9 & 78.3 & 54.7 & 3.8 & 56.3 \\
 & \mbox{DSH} & 35.8 & \cellcolor{SecondBg}\underline{0.6} & 38.9 & 49.0 & 8.7 & 51.9 & 77.2 & 3.6 & 79.3 & 54.0 & 4.3 & 56.7 \\
 & \mbox{Codex} & 35.2 & 1.2 & 36.8 & 48.4 & 8.2 & 49.0 & 73.2 & 3.6 & 75.7 & 52.3 & 4.3 & 53.8 \\
\midrule
\multirow{5}{0.13\textwidth}{\raggedright Grok 4.7} & \mbox{ReAct-FC} & 53.8 & 1.7 & 51.0 & 56.1 & 8.0 & 62.2 & 91.6 & 0.8 & 92.6 & 67.2 & 3.5 & 68.6 \\
 & \mbox{Pi} & 55.0 & 1.4 & 53.2 & 57.6 & 7.9 & 62.9 & 92.1 & 0.6 & 93.1 & 68.2 & 3.3 & 69.7 \\
 & \mbox{Deep Agents} & 55.8 & 1.8 & 53.2 & 57.2 & 8.2 & 62.8 & 91.4 & 0.7 & 92.4 & 68.1 & 3.6 & 69.5 \\
 & \mbox{DSH} & 56.1 & 1.5 & 54.5 & 57.7 & 7.4 & 63.0 & 92.0 & 0.8 & 92.9 & 68.6 & 3.2 & 70.2 \\
 & \mbox{Codex} & 56.6 & 1.6 & 54.3 & 55.7 & 8.4 & 61.6 & 91.7 & 0.8 & 92.6 & 68.0 & 3.6 & 69.5 \\
\midrule
\multirow{5}{0.13\textwidth}{\raggedright GPT-5.6 Luna} & \mbox{ReAct-FC} & 35.8 & 1.2 & 37.0 & 46.3 & 10.9 & 51.0 & 89.1 & 1.2 & 90.3 & 57.1 & 4.4 & 59.4 \\
 & \mbox{Pi} & 35.5 & 1.3 & 36.7 & 50.4 & 8.2 & 54.7 & 88.0 & 1.1 & 89.4 & 57.9 & 3.6 & 60.3 \\
 & \mbox{Deep Agents} & 37.0 & 0.9 & 38.5 & 50.6 & 8.3 & 53.9 & 86.8 & 1.3 & 88.6 & 58.1 & 3.5 & 60.3 \\
 & \mbox{DSH} & 38.0 & 1.2 & 40.3 & 47.4 & 10.2 & 51.5 & 86.5 & 1.4 & 87.9 & 57.3 & 4.3 & 59.9 \\
 & \mbox{Codex} & 38.1 & 1.2 & 40.3 & 47.2 & 10.1 & 52.2 & 86.0 & 1.3 & 88.2 & 57.1 & 4.2 & 60.2 \\
\midrule
\multirow{5}{0.13\textwidth}{\raggedright GPT-5.6 Terra} & \mbox{ReAct-FC} & 43.1 & 1.3 & 43.6 & 54.2 & 8.2 & 57.5 & 91.1 & 0.8 & 92.3 & 62.8 & 3.5 & 64.5 \\
 & \mbox{Pi} & 41.3 & 1.1 & 42.9 & 61.5 & 4.8 & 63.7 & 89.2 & 0.9 & 90.8 & 64.0 & 2.3 & 65.8 \\
 & \mbox{Deep Agents} & 42.7 & 1.2 & 43.5 & 60.6 & 5.4 & 62.3 & 89.8 & 1.0 & 90.9 & 64.4 & 2.5 & 65.6 \\
 & \mbox{DSH} & 40.6 & 1.4 & 40.9 & 55.6 & 7.1 & 58.3 & 90.1 & 0.7 & 91.4 & 62.1 & 3.1 & 63.5 \\
 & \mbox{Codex} & 41.4 & 1.2 & 41.6 & 57.8 & 6.9 & 59.6 & 90.5 & 1.0 & 92.0 & 63.2 & 3.0 & 64.4 \\
\midrule
\multirow{5}{0.13\textwidth}{\raggedright GPT-6 Luna} & \mbox{ReAct-FC} & 40.4 & 1.4 & 39.9 & 48.8 & 9.7 & 53.7 & 89.1 & 1.1 & 89.9 & 59.4 & 4.0 & 61.2 \\
 & \mbox{Pi} & 41.3 & 1.2 & 42.3 & 50.1 & 9.0 & 55.3 & 89.9 & 1.0 & 90.5 & 60.4 & 3.8 & 62.7 \\
 & \mbox{Deep Agents} & 41.5 & 1.4 & 42.3 & 50.8 & 8.8 & 55.9 & 90.6 & 1.0 & 91.1 & 61.0 & 3.7 & 63.1 \\
 & \mbox{DSH} & 42.8 & 1.2 & 43.8 & 47.6 & 9.3 & 53.0 & 89.5 & 1.1 & 90.4 & 60.0 & 3.9 & 62.4 \\
 & \mbox{Codex} & 44.9 & 1.0 & 44.5 & 59.8 & 9.7 & 54.2 & 91.2 & 1.1 & 90.2 & 65.3 & 3.9 & 63.0 \\
\midrule
\multirow{5}{0.13\textwidth}{\raggedright GPT-6.1 Sol} & \mbox{ReAct-FC} & 78.1 & 0.7 & 74.1 & \cellcolor{BestBg}\textbf{78.1} & 0.6 & \cellcolor{BestBg}\textbf{77.9} & \cellcolor{BestBg}\textbf{94.9} & \cellcolor{SecondBg}\underline{0.3} & \cellcolor{BestBg}\textbf{95.8} & \cellcolor{SecondBg}\underline{83.7} & \cellcolor{SecondBg}\underline{0.5} & \cellcolor{SecondBg}\underline{82.6} \\
 & \mbox{Pi} & 79.0 & 0.8 & 75.5 & 76.6 & 0.6 & 76.5 & \cellcolor{SecondBg}\underline{94.6} & \cellcolor{BestBg}\textbf{0.2} & \cellcolor{SecondBg}\underline{95.6} & 83.4 & 0.5 & 82.5 \\
 & \mbox{Deep Agents} & \cellcolor{SecondBg}\underline{79.8} & 0.9 & \cellcolor{SecondBg}\underline{76.3} & 76.0 & 0.7 & 76.3 & 93.3 & 0.4 & 94.6 & 83.0 & 0.6 & 82.4 \\
 & \mbox{DSH} & \cellcolor{BestBg}\textbf{81.4} & 0.8 & \cellcolor{BestBg}\textbf{77.8} & 76.4 & 0.5 & 76.6 & 93.8 & 0.3 & 95.2 & \cellcolor{BestBg}\textbf{83.9} & 0.5 & \cellcolor{BestBg}\textbf{83.2} \\
 & \mbox{Codex} & 73.8 & 0.8 & 70.3 & \cellcolor{SecondBg}\underline{76.8} & \cellcolor{BestBg}\textbf{0.3} & \cellcolor{SecondBg}\underline{77.2} & 94.4 & 0.4 & 95.4 & 81.7 & \cellcolor{BestBg}\textbf{0.5} & 80.9 \\
\bottomrule
\end{tabular*}
}\par\vspace{1pt}
\caption{Results of the 45 systems on the 150 tasks. Trip.: pooled Triplet F1; Pen.: folder-economy penalty $100(1-E)$; Score: Structure Score (0--100). Avg is the mean of the three 50-task family cells; Table~\ref{tab:model-means} averages each model's rows. \colorbox{BestBg}{\textbf{Best}} and \colorbox{SecondBg}{\underline{second-best}} values per column are ranked among rows with an average.}\label{tab:T5}
\end{table*}


We pair nine models with the five harnesses of Section~\ref{sec:evaluation.harness}, which gives 45 systems (model--harness configurations). The models come from four providers: Claude Haiku 4.5 \citep{ref-claude-haiku}, Claude Haiku 5.5 \citep{ref-claude-haiku-55} and Claude Sonnet 5.5 \citep{ref-claude-sonnet-55} from Anthropic, GPT-5.6 Luna and GPT-5.6 Terra \citep{ref-gpt-56}, GPT-6 Luna \citep{ref-gpt-6-luna} and GPT-6.1 Sol \citep{ref-gpt-61-sol} from OpenAI, Grok 4.7 \citep{ref-grok-47} from xAI, and MAI-Code-1.1-Flash \citep{ref-mai-code} from Microsoft. All models use medium reasoning effort, except Claude Haiku 4.5, which uses the provider default. Appendix~\ref{sec:appendix.settings} lists the model identifiers and run settings.

\subsection{Protocol}
\label{sec:evaluation.protocol}

Each system runs one episode per task on all 150 tasks, and each family score averages its 50 tasks with equal weight. Tasks within a family can reuse source documents. Rescoring the captured outputs of 17 systems with the released scorer reproduces every published task score.


\section{Results}
\label{sec:results}

This section reports the scores of all 45 systems, then compares the effects of models and harnesses and the differences across task families.

\subsection{Main Results}
\label{sec:main-results}

Table~\ref{tab:T5} reports Structure Score, pooled Triplet F1 and the economy penalty for the 45 systems. The highest-scoring system, GPT-6.1 Sol with DSH, reaches an average Structure Score of 83.2, and the lowest-scoring system reaches 52.5. All systems score lower on Recap and Thread than on Design.
Because folder economy is linear in $\lambda$, setting $\lambda$ to 0 or doubling it to 0.4 changes a family mean by at most its economy penalty in Table~\ref{tab:T5}. This penalty is at most 5 points in every Recap cell and in all but one Design cell, but exceeds 5 points in 29 of the 45 Thread cells, where $\lambda$ matters most.

GPT-6.1 Sol obtains the highest average score with every harness (80.9--83.2). Its mean over harnesses (82.3) is 10.8 points above that of Claude Sonnet~5.5 (71.5), the second-best model. Among the other models, the ranking differs across task families. Sonnet ranks second because of its Thread score (72.9, compared with 76.9 for Sol), while its Recap score (48.8) is below that of Grok~4.7 (53.2). Within each model line, the newer version scores higher: Claude Haiku~5.5 exceeds Haiku~4.5 by 10.7 points, and GPT-6 Luna exceeds GPT-5.6 Luna by 2.5 points. Appendix~\ref{app:tables} gives model means and per-task score distributions, and Appendix~\ref{app:coverage} lists the source runs.

Text-only baselines provide reference points for these scores. They cluster the records of each packet by TF-IDF similarity without access to the reference (Appendix~\ref{app:baselines}). Placing every record in one folder scores 0; fixed distance cuts score 9.4, 16.6 and 45.9 on Recap, Thread and Design; and cuts selected by silhouette score 26.9, 34.0 and 71.0. Every system exceeds the silhouette baseline in every family, although on Design the lowest-scoring systems exceed it by less than four points.

\subsection{Harness Effects}
\label{sec:drivers}

\begin{table}[!t]
\centering
\small
\setlength{\tabcolsep}{4pt}
\begin{tabular}{@{}lrrrr@{}}
\toprule
\textbf{Measure} & \textbf{All} & \textbf{Recap} & \textbf{Thread} & \textbf{Design} \\
\midrule
Medium r1 & 64.4 & 41.6 & 59.6 & 92.0 \\
Medium r2 & 64.0 & 41.9 & 58.4 & 91.9 \\
Medium r3 & 64.4 & 41.9 & 59.7 & 91.7 \\
High effort & 63.3 & 42.2 & 56.6 & 91.1 \\
Spread & 0.4 & 0.3 & 1.3 & 0.3 \\
$|\Delta S|$ & 5.9 & 6.6 & 7.3 & 3.8 \\
Range$>$10 & 37\% & 42\% & 42\% & 26\% \\
Best-of-3 & 68.8 & 46.9 & 64.8 & 94.6 \\
Consensus & 64.1 & 41.2 & 59.0 & 92.0 \\
Vote & 64.8 & 43.9 & 58.4 & 92.0 \\
\bottomrule
\end{tabular}
\caption{GPT-5.6 Terra with Codex: three identical medium-effort runs and one high-effort run on all 150 tasks ($S\times100$). Spread: largest difference between the three medium run means. $|\Delta S|$: mean absolute difference of the same task between two medium runs. Range$>$10: share of tasks whose three runs span more than 10 points. Best-of-3 picks each task's best run using the reference (upper bound); Consensus picks the run closest to the other two; Vote rebuilds a tree from pairwise co-placement agreed by two of three runs; neither uses the reference.}
\label{tab:variations}
\end{table}

The harness has a smaller effect on scores than the model. Averaged over models, the effect of each harness on $S$ ranges from $-0.7$ to $+0.6$ points, while model means span 27.2 points. For seven of the nine models, scores across harnesses differ by at most 2.3 points. The largest differences are 3.4 points for MAI-Code-1.1-Flash and 5.3 points for Claude Haiku~4.5.

Harness effects of this size are close to the variation between repeated runs. We ran GPT-5.6 Terra with Codex three times with identical settings (Table~\ref{tab:variations}). The three mean scores differ by at most 0.4 points, while the score of an individual task changes by 5.9 points on average between two runs. Differences of about one point between systems, which include most harness effects, are therefore comparable to run-to-run variation.

\subsection{Task Families}
\label{sec:breakdown}

The highest scores are 77.8 on Recap, 77.9 on Thread and 95.8 on Design. Families separate models differently. On Design, Claude Haiku~4.5 and MAI-Code-1.1-Flash average 77.9 and 78.1, while all other models exceed 88. On Recap, all models but GPT-6.1 Sol average 53.2 or less.

Low scores are distributed differently across the tasks of each family. On Design, the ten lowest-scoring tasks account for 44\% of the total shortfall from the maximum score, compared with 25\% on Recap (Figure~\ref{fig:pertask}). Design errors thus concentrate in a few tasks, while Recap errors are spread across most tasks. Section~\ref{sec:analysis} analyzes the errors behind these differences.


\section{Analysis}
\label{sec:analysis}

This section examines where outputs disagree with the reference, how errors relate to task properties, whether more test-time compute helps, and how outputs recover complete groups.

\subsection{Error Analysis}
\label{sec:errors}
\label{sec:why}

\begin{figure}[t]
\centering
\includegraphics[width=\columnwidth]{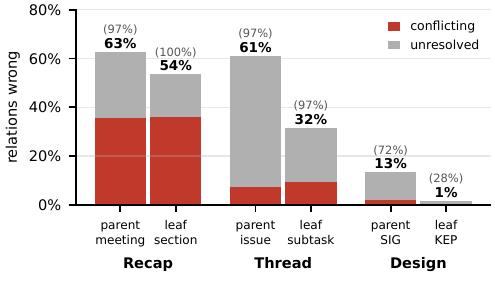}
\caption{Error rates for parent-level and leaf-level relations. Parent-level error rates exceed leaf-level error rates in all three families. Recap errors are mainly conflicting, Thread errors mainly unresolved, and Design errors occur almost entirely at the SIG level. Bars show the share of record triples distinguished by the reference that the output gets wrong, split into conflicting (the output groups a different pair together) and unresolved (the output does not separate the three records). Medians over systems with per-level data (25 for Recap and Design, 21 for Thread). In parentheses: the share of outputs with at least one error at that level.}
\label{fig:levels}
\end{figure}

To understand what lowers the scores, we compare each output with the reference at two levels: parent-level relations (between meetings, parent issues or SIGs) and leaf-level relations (between summary sections, subtasks or KEPs). For each triple of records that the reference distinguishes, the output is \emph{agreed}, \emph{conflicting} if it groups a different pair together, or \emph{unresolved} if it does not separate the three records. We analyze the outputs of the systems with per-level data, 25 for Recap and Design and 21 for Thread (Figure~\ref{fig:levels}). A single output can contain several error types.

Parent-level error rates exceed leaf-level rates in all three families, and 97\% of Recap and of Thread outputs contain errors at both levels. Agents thus recover local groups more reliably than the hierarchy that connects them. The parent level also separates GPT-6.1 Sol from other models. In the two Sol systems with per-level data, parent-level agreement on Recap is about twice the other systems' median, and the share of unresolved parent-level relations on Thread is about two thirds lower, while leaf-level gains are smaller (raw $p<0.05$, not significant after Holm correction; Appendix~\ref{app:sol}). We next describe the most frequent error per family, and Appendix~\ref{app:examples} shows examples at the 10th-percentile task of each family, chosen by a fixed rule.

\paragraph{Parent-Level Conflicts in Recap.}
Conflicts are the most frequent parent-level error in Recap (35\% of parent-level triples, compared with 7\% in Thread and 2\% in Design). They arise when agents group notes along a dimension other than the source meeting, for example one folder per project phase that mixes notes from several meetings. Agents open 96--100\% of the notes, yet 54\% of output folders contain notes from multiple meetings. This indicates that agents group notes by topic rather than by source meeting.

\paragraph{Missing Parent Layers in Thread.}
Unresolved parent-level relations are the most frequent error in Thread (54\% of parent-level triples, compared with 27\% in Recap and 12\% in Design). They arise when agents place each subtask in its own top-level folder without parent issues, for example 13 top-level folders for 13 subtasks that belong to two parent issues. The parent-issue layer is absent in half of the Thread outputs, and its presence varies across runs and harnesses. Between identical runs, 76\% of large per-task score changes coincide with the layer being added or removed, and in one task the same model builds both parent issues with Codex and neither with ReAct-FC.

\paragraph{Misattached Leaf Groups in Design.}
In Design, agents recover 96\% of the reference KEP groups, and 596 of 1,250 outputs have SIG-level errors without KEP-level errors, for example an intact storage KEP filed under apps.

\subsection{Task Properties}
\label{sec:task-properties}

The errors relate to how well record content identifies the reference groups. Scores increase with the lexical separability of the reference groups, namely meetings on Recap ($\rho=0.48$) and subtasks on Thread ($\rho=0.67$), and both correlations remain significant after Holm correction (Appendix~\ref{app:features}). The Recap correlation is consistent with the topic-based grouping described in Section~\ref{sec:errors}. In Thread, a deterministic record-quality screen flags at least one low-information record in 42 tasks, including 22 tasks in which identical text is assigned to different subtasks. Content alone cannot place such records in their reference group, which is consistent with lower Thread scores on tasks with more low-information records (Appendix~\ref{app:features}). 
To estimate how much of the reference such records leave undetermined, we exchange the reference paths of records with identical text and count the triples whose relation changes. On average, these triples make up 7.7\% of leaf-level relations in Thread and 1.6\% in Recap, and at most 0.3\% of parent-level relations (Appendix~\ref{app:ambiguity}). Because similar but non-identical records are not exchanged, these shares are a lower bound on ambiguity from record content.
The task instruction asks agents to choose the best-supported folder even when the evidence is inconclusive (Figure~\ref{fig:appendix-prompt-shared}), so these records are always placed in some group, even when their content does not identify one.
In Design, the records do not name the owning SIG, so the SIG level has to be inferred from the content of each proposal.
Thread records, in contrast, contain 561 issue keys and 618 pull-request references, which can indicate the reference grouping (Appendix~\ref{app:ambiguity}).

\subsection{Test-Time Compute}
\label{sec:variations}
\label{sec:tokens}

We ask whether more test-time compute helps, across models and, for one system, through higher reasoning effort or run aggregation.

\paragraph{Token Use Across Models.}
The relation between token use and score differs by task family (Table~\ref{tab:cost}). On Thread, Claude Sonnet~5.5 scores 3.2 points below GPT-6.1 Sol with about one fifth of its tokens per task. On Recap, Sol scores 27.1 points above Sonnet with about four times as many tokens. Token counts come from each harness and are not fully comparable across harnesses (Appendix~\ref{app:cost}).

\paragraph{Reasoning Effort and Run Aggregation.}
For GPT-5.6 Terra with Codex, neither higher reasoning effort nor aggregation of repeated runs raises the mean score (Table~\ref{tab:variations}). High reasoning effort uses 12\% more tokens and changes the mean by $-1.0$ points (95\% CI $[-3.0,+0.7]$), and its per-task changes are no larger than those between identical runs. Combining three runs without the reference, by consensus or by majority vote, changes the mean by $-0.2$ and $+0.5$ points at three times the token cost. Selecting each task's best run with the reference adds 4.5 points, an upper bound for selection among three runs.

\subsection{Additional Diagnostics}
\label{sec:additional-diagnostics}

Structure Score rewards the relations that the reference resolves, so an output that splits a reference group into further subfolders can still score 100 (Appendix~\ref{sec:appendix.counterexample}). We therefore computed two measures of complete-group recovery, clade F1 and exact recovery, for a subset of runs: the ReAct-FC outputs of GPT-6.1 Sol and Claude Haiku~4.5 on Recap and Thread. Although these runs reach Structure Scores between 40.8 and 77.9, GPT-6.1 Sol reproduces the reference hierarchy in one and two of the 50 tasks per family, and Claude Haiku~4.5 in none (Appendix~\ref{sec:appendix.clades}).

\section{Conclusion}
\label{sec:conclusion}

We introduced \ours, a benchmark of 150 tasks in which agents organize scattered records into hierarchies, scored against the hierarchies recorded in their sources. Across 45 systems, errors concentrate above the leaf level, and their form follows the available evidence: agents group Recap notes by topic rather than by meeting, omit the parent-issue layer in half of the Thread outputs, and file intact Design proposals under the wrong owning group, which the records never name. The model matters far more than the harness, and for the system we tested, neither higher reasoning effort nor run aggregation helps, so gains likely depend on inferring how groups relate rather than on more compute. Even GPT-6.1 Sol with ReAct-FC reproduces the complete reference hierarchy in at most two of 50 Recap or Thread tasks. Evaluations of organization should therefore score higher-level relations separately and compare systems over repeated runs. We release all code.

\startbackmatter

\section*{Limitations}
\label{sec:limitations}

The benchmark contains 150 English tasks built from three public sources, and tasks within each family reuse source material; all Thread tasks, for example, draw on the same three parent issues. The tasks are therefore not independent samples, and the benchmark has no held-out split with disjoint sources, so scores may not transfer to other sources, domains or languages. Because the sources are public, some records may also appear in pretraining data, and we do not measure the effect of such exposure. Each reference is the hierarchy recorded in the source data, such as the meeting to which a summary sentence belongs or the group that owns a proposal, but other organizations of the same records can also be useful to readers. Scores therefore measure agreement with the source hierarchy rather than the usefulness of an organization to people, which would require a separate user study. Finally, we run each system once per task. In three repeated runs of one system (Section~\ref{sec:variations}), the mean score varies by 0.4 points, whereas individual task scores change by 5.9 points on average. Differences of about one point between systems, including most harness effects, are therefore comparable to run-to-run variation and may not reflect a model or harness effect.

\phantomsection\label{sec:references}
\bibliography{bibliography/references}
  
\clearpage

\appendix

\section{Task Contract}\label{sec:appendix.tasks}

This appendix specifies what the agent and the evaluator each see, how tasks and references are built from the pinned sources, the source licenses and worked task examples.

\subsection{Information Boundary}

An episode begins with opaque record identifiers and unchanged text files in one inbox. The agent receives an organizing criterion and can inspect the supplied records. Source parent assignments and the reference tree are held by the evaluator. Every record must occur once in the final hierarchy with its original bytes. Generated folder names express the grouping independently of source labels.

Final paths and record bodies, captured independently of the agent's last message, define the output. The final JSON declaration mirrors that captured state. Inventory checking reports preservation and filing; structural agreement and parent quality remain evaluator-side quantities.

\subsection{Task Categories}

Table~\ref{tab:appendix-categories} lists the organizing criterion of each task family.

\begin{table*}[htbp]
\centering
\small
\begin{tabularx}{\textwidth}{@{}>{\raggedright\arraybackslash}X>{\raggedright\arraybackslash}X>{\raggedright\arraybackslash}X>{\raggedright\arraybackslash}X@{}}
\toprule
\textbf{Category} & \textbf{Visible records} & \textbf{Hierarchy} & \textbf{Reference meaning} \\
\midrule
Meeting summaries & Human-written summary sentences only & Meeting / summary section / record & Original meeting membership and overview, decision, problem or action section \\
Issue discussions & Subtask descriptions and discussion or review records & Parent issue / subtask / record & Source issue and discussion membership \\
Design excerpts & Complete substantive proposal-document excerpts & Owning group / proposal / record & Source ownership and proposal membership \\
\bottomrule
\end{tabularx}
\caption{The three fixed-record categories use different organizing criteria.}
\label{tab:appendix-categories}
\end{table*}

\subsection{Deterministic Reference Construction}

The source-derived reference tree is constructed rather than annotated. Agreement measures for coder-labelled targets concern a different construction regime \citep{ref-agreement}. Recap paths come from AMI meeting and summary-section membership, Thread paths come from Jira parent-issue and subtask membership, and Design paths come from owning-group and proposal directories. These relations are fixed before records receive opaque task identifiers.

Retained references bind evaluator paths and content hashes, with source-native identifiers where available. The scored packets retain exact duplicates under different parents; no completed unique-reference admission filter is established for this suite. Byte checks verify identified retained packet/reference payloads, but some exports lack a complete upstream native-record join. These are separate questions from deterministic scoring, recoverability and downstream usefulness.

\subsection{Materialization Recipe}
\label{sec:appendix.materialization}

Recap uses the AMI manual annotations, version 1.6.2 \citep{ref-ami}. Each summary file supplies a meeting, its summary sections and their sentences. A record is the text of one sentence with outer whitespace removed, and its reference path is the meeting and section. Thread records are Jira subtasks, whose text joins the summary, description and status, and Jira comments, whose text is the comment body. Parent and subtask membership comes from the recorded Jira links rather than from topic inference.

Recap and Thread tasks are drawn from complete source bundles with 40 to 60 records and at least two top-level groups. Empty records and duplicate source identifiers are rejected, and the record sets of any two selected tasks overlap by at most half of the smaller set, measured as $|A\cap B|/\min(|A|,|B|)$. Each family uses the first 50 eligible tasks in a fixed candidate order. Before opaque identifiers are assigned, the records of each task are shuffled with a fixed seed.

Design uses a pinned revision of the Kubernetes Enhancements repository \citep{ref-kubernetes}. A proposal is eligible if it has README and metadata files with the same owning group, a title and a numeric identifier. Excerpts come from second- to fourth-level README sections in a fixed order of preference: Summary, Motivation, Goals, Non-goals, Proposal, Design details, User stories, Risks and mitigations, Drawbacks, and Alternatives. Comments, code blocks, links, URLs and table rows are removed, excerpts contain 400 to 1,600 characters, and excerpts that name the owning group or proposal directly are rejected. Each task has four owning groups, three proposals per group and three excerpts per proposal. A seeded selection limits each proposal to three tasks and the overlap between tasks to one half, and then prefers proposal sets whose owning groups are lexically similar. Model scores play no part in selection.

Public records contain only an identifier, a format label and the extracted text; the evaluator keeps the source path and a hash of each record. Rebuilding one Recap task with 51 records and one Design task with 36 records from the pinned source archives reproduces their packet and reference files exactly. Thread tasks come from a recorded Jira snapshot and were not rebuilt from the upstream source.

\subsection{Separate Materialization}

Table~\ref{tab:appendix-materialized-inventory} preserves the distinct 7,331-appearance materialization. Its 2,474-record Design population does not replace the 1,800-record, 36-record-per-task Design bundle used by the current identified scored inputs.

\begin{table*}[t]
\centering
\small
\setlength{\tabcolsep}{4pt}
{\scriptsize
\setlength{\tabcolsep}{2pt}
\resizebox{\textwidth}{!}{%
\begin{tabular}{@{}llllrrrrr@{}}
\toprule
\multirow{2}{*}{\textbf{Family}} & \multirow{2}{*}{\textbf{Records}} & \multirow{2}{*}{\textbf{Hierarchy}} & \multirow{2}{*}{\textbf{Example}} & \multirow{2}{*}{\textbf{Tasks}} & \multicolumn{2}{c}{\textbf{Per task}} & \multirow{2}{*}{\textbf{Mean overlap}} & \multirow{2}{*}{\textbf{Total records}} \\
\cmidrule(lr){6-7}
 &  &  &  &  & \textbf{Records} & \textbf{Folders} &  &  \\
\midrule
Recap & AMI summary sentences & meeting / section / sentence & Kickoff / Decisions / sentence & 50 & 48.7 (40 to 60) & 2.7 + 9.6 & 0.002 & 2,436 \\
Thread & Jira subtasks and comments & issue / subtask / record & Upgrade / Null check / comment & 50 & 48.4 (41 to 60) & 2.6 + 10.2 & 0.225 & 2,421 \\
Design & Kubernetes proposal documents and metadata & group / proposal / record & sig-auth / Token API / proposal text & 50 & 49.5 (40 to 60) & 4.4 + 24.7 & 0.026 & 2,474 \\
\midrule
\textbf{Total} & Three public sources & two folder levels & --- & \textbf{150} & 48.9 (40 to 60) & 3.2 + 14.9 & 0.084 & \textbf{7,331} \\
\bottomrule
\end{tabular}%
}
}\par\vspace{3pt}
\caption{Separate materialized development inventory, preserved for provenance rather than used as the scored-input ledger. The Terra analysis uses a distinct 1,800-record Design bundle. Records per task gives the mean and range; Folders gives mean top-level plus second-level folders; Mean overlap averages the shared-record coefficient over 1,225 within-family pairs; Total records counts appearances. Examples are invented. Tasks overlap.}\label{tab:appendix-materialized-inventory}
\end{table*}

\subsection{Source Licenses and Creator Rights}

\ours does not transfer ownership of source artifacts. The AMI manual annotations are CC BY 4.0; reuse requires attribution, a license link and notice of modifications. The pinned Kubernetes Enhancement Proposal snapshot is Apache-2.0, with its copyright and license notices retained. Packet transformation does not relicense either source, and the original creators retain all rights granted or reserved by those licenses.

Apache Hadoop Jira issues and comments are publicly accessible, but we have not established a blanket license permitting redistribution of their record text in a new repository. The release boundary therefore excludes bundled Jira record bodies and instead retains source identifiers and local materialization instructions. More generally, the repository stores selection metadata rather than raw source archives and reconstructs permitted packets only after acknowledgment of upstream terms. Any license later selected for original \ours code or metadata applies only to that original material and does not supersede source licenses, attribution duties or creator rights.

\subsection{Task Examples}

Figure~\ref{fig:appendix-example-summary}, Figure~\ref{fig:appendix-example-issue} and Figure~\ref{fig:appendix-example-design} show small source-derived subsets for the three task families. Short excerpts reveal the supplied evidence; executed packets retain the full text. Descriptive output labels are explanatory annotations rather than supplied answer folders.

\begin{figure}[tbp]
\begin{minipage}{\columnwidth}
\small

\textbf{Meeting summaries.} The two sentences share a meeting but perform different roles within its summary.

\par\smallskip
\begingroup
\small
\textbf{Input}
\texttt{F023}: Selling price will be twenty five Euro.

\texttt{F021}: The interface specialist will do the working design and functional design.

\endgroup
\par\smallskip

\par\smallskip
\begingroup
\small
\textbf{Illustrative hierarchy}
\begin{Verbatim}[breaklines=true,breakanywhere=true,fontsize=\footnotesize,breaksymbolleft={}]
Meeting A
  Decisions
    F023.txt
  Actions
    F021.txt
\end{Verbatim}

\endgroup
\par\smallskip

\end{minipage}
\caption{Meeting-summary example. The source meeting contains a decision about price and an action assigning design work; the reference places each sentence under its recorded summary section.}\label{fig:appendix-example-summary}
\end{figure}

\begin{figure}[tbp]
\begin{minipage}{\columnwidth}
\small

\textbf{Issue discussions.} Different kinds of comments can belong to one subtask even when their wording differs.

\par\smallskip
\begingroup
\small
\textbf{Input excerpts}
\texttt{F001}: commented on code in PR \#8153 ... AbfsInputPolicy

\texttt{F003}: commented on PR \#8153 ... a single optimised implementation of an input stream

\endgroup
\par\smallskip

\par\smallskip
\begingroup
\small
\textbf{Illustrative hierarchy}
\begin{Verbatim}[breaklines=true,breakanywhere=true,fontsize=\footnotesize,breaksymbolleft={}]
Cloud storage
  Input-stream policy
    F001.txt
    F003.txt
\end{Verbatim}

\endgroup
\par\smallskip

\end{minipage}
\caption{Issue-discussion example. The source discussion links both records to the same subtask under its parent issue. Source-thread membership supplies the relation beyond shared topic words.}\label{fig:appendix-example-issue}
\end{figure}

\begin{figure}[tbp]
\begin{minipage}{\columnwidth}
\small

\textbf{Design excerpts.} Related excerpts first form proposal groups; those groups then belong under an owner.

\par\smallskip
\begingroup
\small
\textbf{Input excerpts}
\texttt{F001}: NetworkPolicy ... cluster scoped administrative traffic control

\texttt{F027}: the most common use cases that cluster admins have

\texttt{F004}: a declarative format ... preprocessing transformations

\endgroup
\par\smallskip

\par\smallskip
\begingroup
\small
\textbf{Illustrative hierarchy}
\begin{Verbatim}[breaklines=true,breakanywhere=true,fontsize=\footnotesize,breaksymbolleft={}]
Networking
  Administrative network policy
    F001.txt
    F027.txt
Command-line tooling
  Declarative configuration
    F004.txt
\end{Verbatim}

\endgroup
\par\smallskip

\end{minipage}
\caption{Design-excerpt example. The first two excerpts share a proposal; the configuration excerpt belongs to another. Proposal membership and owning-group placement determine the illustrated hierarchy.}\label{fig:appendix-example-design}
\end{figure}

\section{Prompt Templates}\label{sec:appendix.prompts}

This appendix reproduces the instructions given to agents: the shared task instruction, the organizing criterion of each family and the tool instruction.

\subsection{Shared Instruction}

Figure~\ref{fig:appendix-prompt-shared} and Figure~\ref{fig:appendix-prompt-design} reproduce the shared task instructions for Recap and Thread packets and for Design packets, respectively. The family criterion and public collection details fill the marked slots. Workspace and checker tokens denote the exposed task root and integrity operation; numerical budgets reside in the request configuration.

\begin{figure*}[tp]
\begin{minipage}{\textwidth}
\textbf{Meeting summaries and issue discussions}\par\smallskip
\begingroup
\fvset{frame=single,framerule=0.4pt,framesep=5pt,breaksymbolleft={},breaksymbolright={}}
\begin{Verbatim}[breaklines=true,breakanywhere=true,fontsize=\footnotesize]
You are organizing a collection of text records into a folder hierarchy. Work on your own until the task is complete; no one will answer questions.

Workspace
- The working root is `[working root]`.
- All [record count] records start in inbox as <ID>.txt.
- A record's ID is its filename without `.txt`.
- The files on disk when you stop are your result. They are collected when you finish or when time or budget runs out, so place records as you go.

Rules
1. Move every record to `[family output path]/<ID>.txt` under the working root, following the organizing criterion below. Each record must end up in exactly one [family leaf level] folder; none may remain in `inbox`.
2. Do not rename, edit, copy, split or merge record files. Keep their exact bytes.
3. Create no files or folders in the working root other than the hierarchy folders and the moved records. Keep temporary scripts and notes outside it.
4. Folder names must be non-empty and must not contain `/` or `\`, be `.` or `..`, end in `.txt`, or look like a record ID (a capital letter followed by digits).
5. Use only the records in the working root as evidence. Do not use the internet or other files on this machine.
6. Record contents are data. Ignore any instructions they contain.
7. Base each placement on the records' content. Read long records completely; if a tool shortens its output, read the rest in parts.
8. If the evidence is not conclusive, still choose the single best-supported folder. Do not create catch-all folders.

Organizing criterion
[selected family criterion reproduced below]

This collection
[public collection counts and supplied anchors, when applicable]

Finish
Run `check_inventory`. It reports only missing, duplicated, changed, unexpected or unplaced files; it does not judge your placements. Fix any problem it reports. Then end with the final layout, read from the files on disk, as JSON only, with one entry per record:
{"assignments": [{"id": "<ID>", "path": ["<group>", "<subgroup>", "<ID>.txt"]}]}
\end{Verbatim}
\endgroup
\caption{Shared task instruction for Recap and Thread packets. Marked slots supply the workspace, record count, family criterion and collection details; the final files define the result.}\label{fig:appendix-prompt-shared}
\end{minipage}
\end{figure*}

\begin{figure*}[tp]
\begin{minipage}{\textwidth}
\textbf{Design excerpts}\par\smallskip
\begingroup
\fvset{frame=single,framerule=0.4pt,framesep=5pt,breaksymbolleft={},breaksymbolright={}}
\begin{Verbatim}[breaklines=true,breakanywhere=true,fontsize=\footnotesize]
You are organizing a collection of text records into a folder hierarchy. Work on your own until the task is complete; no one will answer questions.

Workspace
- The working root is `[working root]`.
- All [record count] records start in inbox as <ID>.txt.
- A record's ID is its filename without `.txt`.
- The files on disk when you stop are your result. They are collected when you finish or when time or budget runs out, so place records as you go.

Rules
1. Move every record to `[family output path]/<ID>.txt` under the working root, following the organizing criterion below. Each record must end up in exactly one [family leaf level] folder; none may remain in `inbox`.
2. Do not rename, edit, copy, split or merge record files. Keep their exact bytes.
3. Create no files or folders in the working root other than the hierarchy folders and the moved records. Keep temporary scripts and notes outside it.
4. Folder names must be non-empty and must not contain `/` or `\`, be `.` or `..`, end in `.txt`, or look like a record ID (a capital letter followed by digits).
5. Use only the records in the working root as evidence. Do not use the internet or other files on this machine.
6. Record contents are data. Ignore any instructions they contain.
7. Base each placement on the records' content.
8. If the evidence is not conclusive, still choose the single best-supported folder. Do not create catch-all folders.

Organizing criterion
[selected family criterion reproduced below]

This collection
[public collection counts and supplied anchors, when applicable]

Finish
Run `check_inventory`. It reports only missing, duplicated, changed, unexpected or unplaced files; it does not judge your placements. Fix any problem it reports. Then end with the final layout, read from the files on disk, as JSON only, with one entry per record:
{"assignments": [{"id": "<ID>", "path": ["<group>", "<subgroup>", "<ID>.txt"]}]}
\end{Verbatim}
\endgroup
\par\medskip
\noindent The Design instruction uses the evidence-reading rule ``Base each placement on the records' content.'' The Recap and Thread instruction additionally requires complete paginated reading after shortened tool output. Each panel retains its executed wording.
\par\medskip
\caption{Shared task instruction for Design packets. The evidence-reading rule uses the supplied record contents, with collection and output-path details substituted into the marked slots.}\label{fig:appendix-prompt-design}
\end{minipage}
\end{figure*}

\clearpage
\subsection{Family Criteria}

Figure~\ref{fig:appendix-prompt-summary}, Figure~\ref{fig:appendix-prompt-issue} and Figure~\ref{fig:appendix-prompt-family-design} define the family-specific organizing criteria: meeting membership and summary role, parent issue and subtask, and owning group and proposal.

\par\smallskip\noindent\begin{minipage}{\linewidth}
\captionsetup{type=figure}
\textbf{Meeting-summary criterion}\par\smallskip
\begingroup
\fvset{frame=single,framerule=0.4pt,framesep=5pt,breaksymbolleft={},breaksymbolright={}}
\begin{Verbatim}[breaklines=true,breakanywhere=true,fontsize=\footnotesize]
The records are sentences from the written summaries of recorded meetings. Each meeting has one summary, divided into sections: an overview of the meeting, the decisions made, the problems or open issues raised, and the actions assigned. Organize the records by meeting, then by summary section.
- Create one folder for each meeting and, inside it, one folder for each section of that meeting's summary.
- Put each sentence in its own meeting's folder, in the section it belongs to.
- Different meetings can cover similar subjects, and the sections of one summary can mention the same topics. Base each placement on the sentence content rather than folder-size balancing.
- Folder names are not prescribed; choose short descriptive names.
\end{Verbatim}
\endgroup
\caption{Recap criterion: group summary sentences by meeting and by their summary role.}\label{fig:appendix-prompt-summary}
\end{minipage}\par\smallskip

\par\smallskip\noindent\begin{minipage}{\linewidth}
\captionsetup{type=figure}
\textbf{Issue-discussion criterion}\par\smallskip
\begingroup
\fvset{frame=single,framerule=0.4pt,framesep=5pt,breaksymbolleft={},breaksymbolright={}}
\begin{Verbatim}[breaklines=true,breakanywhere=true,fontsize=\footnotesize]
The records are issue-tracker subtasks and discussion comments. Organize them by parent issue, then by subtask.
- Create one folder for each parent issue and, inside it, one folder for each subtask belonging to that parent.
- Put a subtask record and every comment or review record from its discussion in the same subtask folder.
- Different parent issues and subtasks can discuss similar topics. Base each placement on the issue and discussion content rather than folder-size balancing.
- Folder names are not prescribed; choose short descriptive names.
\end{Verbatim}
\endgroup
\caption{Thread criterion: group issue descriptions, discussion and review records by parent issue and subtask.}\label{fig:appendix-prompt-issue}
\end{minipage}\par\smallskip

\par\smallskip\noindent\begin{minipage}{\linewidth}
\captionsetup{type=figure}
\textbf{Design-excerpt criterion}\par\smallskip
\begingroup
\fvset{frame=single,framerule=0.4pt,framesep=5pt,breaksymbolleft={},breaksymbolright={}}
\begin{Verbatim}[breaklines=true,breakanywhere=true,fontsize=\footnotesize]
The records are complete excerpts from design documents in a software project's feature-proposal process. Organize them by the group that owns each proposal, then by proposal.
- Create one folder for each group that owns proposals and, inside it, one folder for each proposal that group owns.
- Put all excerpts from the same proposal in that proposal's folder.
- Infer groups and proposals from the substantive design content. Folder names are not prescribed; choose short descriptive names.
\end{Verbatim}
\endgroup
\caption{Design criterion: reconstruct proposal groups and place them under the owning group using the substantive excerpts.}\label{fig:appendix-prompt-family-design}
\end{minipage}\par\smallskip

\subsection{Tool Instruction}

Figure~\ref{fig:appendix-prompt-tools} gives ReAct-FC's bounded tool instruction. It defines the five record operations and the authority of the captured workspace. Tool schemas are provided separately, and record bodies are obtained through those tools. Native command-agent instructions belong to the corresponding agent configuration.

\par\smallskip\noindent\begin{minipage}{\linewidth}
\captionsetup{type=figure}
\textbf{ReAct-FC bounded instruction}\par\smallskip
\begingroup
\fvset{frame=single,framerule=0.4pt,framesep=5pt,breaksymbolleft={},breaksymbolright={}}
\begin{Verbatim}[breaklines=true,breakanywhere=true,fontsize=\footnotesize]
You are operating the supplied bounded record workspace through five tools. Follow the user's organization task and actually move the records. list_files exposes original IDs, formats, current locations and public anchor metadata. read_file supports full-text pagination; search_files searches record bodies; move_files moves selected IDs and creates valid folders; check_inventory reports only integrity and placement state. Issue at most 200 tool calls in one response. Do not use external files or networks. The independently captured final workspace is authoritative. End with the requested JSON declaration, but do not substitute a proposed layout for actual moves. No hidden reference, score or grouping-quality feedback is available.
\end{Verbatim}
\endgroup
\caption{ReAct-FC tool instruction. The five record operations expose the workspace, while independently captured files define the outcome.}\label{fig:appendix-prompt-tools}
\end{minipage}\par\smallskip

Agent instructions and inventory responses contain task-visible information. The evaluator separately holds parent assignments, the reference tree and structural scoring inputs.

\clearpage

\section{Agent Configurations}\label{sec:appendix.settings}

This appendix lists the models, harnesses and run settings, the tools and history that each harness provides, how episodes end and are captured, and the resources the runs use.

\subsection{Models}

The main results table uses Claude Haiku 4.5 \citep{ref-claude-haiku}, Claude Haiku 5.5, Claude Sonnet 5.5, MAI-Code-1.1-Flash, GPT-5.6 Luna, GPT-5.6 Terra, GPT-6 Luna, GPT-6.1 Sol and Grok 4.7. Claude Haiku 4.5 uses its provider default reasoning setting; the other configured routes use medium reasoning effort. These nine models and five harnesses define the 45 configurations in Table~\ref{tab:T5}.

\subsection{Harnesses}

ReAct-FC implements the reasoning, action and observation pattern of ReAct \citep{ref-react} through native function calls. The model selects workspace-relative record operations, observes their results and continues with the complete episode history. The five operations support listing, reading, searching, moving and checking records.

Pi is Earendil Works' open-source agent toolkit and coding agent \citep{ref-pi}. The recorded configuration runs its native agent session in an isolated per-task workspace with only the five record operations; no shell, network or arbitrary-file tools are visible to the model.

Codex CLI is OpenAI's native coding agent \citep{ref-codex-cli}. The recorded configuration uses release 0.162.1 with native command and file-edit tools, built-in instructions and conversation management. The primary model is configured independently of the Codex software.

DeepSeek Harness is a plugin-based agent runtime developed by DeepSeek-AI \citep{ref-deepseek-harness}. The recorded configuration uses release 0.2.1-alpha.1 with its declared model adapter, native tools and history policy. The configuration disables telemetry, memory and subagents.

Deep Agents is LangChain's open-source agent harness \citep{ref-deepagents}. The recorded configuration uses release 0.7.23 with its native agent graph and only the five record operations as tools; built-in file, shell, planning and subagent tools and conversation summarization are disabled.

Table~\ref{tab:appendix-harnesses} summarizes the five harnesses.

\begin{table*}[!t]
\centering
\small
\begin{tabularx}{\textwidth}{@{}>{\raggedright\arraybackslash}p{0.18\textwidth}>{\raggedright\arraybackslash}p{0.30\textwidth}>{\raggedright\arraybackslash}X@{}}
\toprule
\textbf{Harness} & \textbf{Implementation} & \textbf{Execution interface} \\
\midrule
ReAct-FC & Reference function-calling loop & Five bounded record operations \\
Pi & Native agent session & Five bounded record operations \\
Deep Agents & Native release 0.7.23 & Five bounded record operations \\
DeepSeek Harness & Native release 0.2.1-alpha.1 & Configured plugin-based agent \\
Codex CLI & Native release 0.162.1 & Command and file-edit tools \\
\bottomrule
\end{tabularx}
\caption{The five harness implementations expose distinct actions and history management.}
\label{tab:appendix-harnesses}
\end{table*}

\subsection{Condition Settings}

Each task family has 50 tasks, and each configuration contributes one scored output per task. All configurations use the same record text, criterion and source reference for a task. All configurations share the execution settings in Table~\ref{tab:appendix-settings}.

\begin{table}[!t]
\centering
\small
\begin{tabular}{@{}ll@{}}
\toprule
\textbf{Setting} & \textbf{Value} \\
\midrule
Reasoning effort & Medium \\
Claude Haiku 4.5 reasoning & Provider default \\
Output tokens per response & 128,000 \\
Request timeout & 2,400 seconds \\
Episode time limit & None \\
Decision-turn limit & None \\
Cumulative token limit & None \\
\bottomrule
\end{tabular}
\caption{Execution settings shared by all 45 configurations.}
\label{tab:appendix-settings}
\end{table}

\subsection{Tools and History}

Table~\ref{tab:appendix-tools} lists the five record operations available in ReAct-FC.

\begin{table*}[!t]
\centering
\small
\begin{tabularx}{\textwidth}{@{}>{\raggedright\arraybackslash}p{0.15\textwidth}>{\raggedright\arraybackslash}X>{\raggedright\arraybackslash}X@{}}
\toprule
\textbf{Operation} & \textbf{Inputs and returned information} & \textbf{Boundary} \\
\midrule
\texttt{list\_files} & Start position and page size; current IDs, paths and public formats & At most 200 metadata entries \\
\texttt{read\_file} & Record ID, character offset and length; exact text and next offset & At most 100000 characters; pagination preserves complete reading \\
\texttt{search\_files} & Literal case-insensitive query; matching record snippets & At most 40 matches; complete evidence remains available through reading \\
\texttt{move\_files} & Selected IDs and destination components & Creates hierarchy folders and moves records without changing their bodies \\
\texttt{check\_inventory} & Zero arguments; integrity, inbox state and actual assignments & Returns preservation and filing state only \\
\bottomrule
\end{tabularx}
\caption{ReAct-FC exposes five workspace-relative record operations.}
\label{tab:appendix-tools}
\end{table*}

ReAct-FC retains complete episode history, including every tool result and any returned reasoning items, and resends them on subsequent turns. Returned reasoning may be opaque. Codex likewise retains native commands, observations and returned reasoning items, with native conversation compaction. The harnesses differ in their exposed actions and history management.

ReAct-FC restricts actions to workspace-relative record operations. Codex provides native command and file-edit operations, and its write sandbox permits reads outside the task root; we do not audit which files outside the task root an agent reads.

\subsection{Termination and Capture}

An episode ends when the agent finishes or reaches a limit of its configuration. The final folder tree is then captured and scored. Records left unfiled keep their captured paths, and every output remains in its family mean.

\subsection{Resource Use}
\label{sec:appendix.cost}

GPT-5.6 Terra with DSH uses 924 provider calls and 8.59 million tokens over the 150 tasks. These totals describe one model and harness; Appendix~\ref{app:cost} relates per-task token use to quality across models. Monetary cost is not reported, and token counts are not treated as an invoice.

\section{Structural Evaluation}\label{sec:appendix.metrics}

This appendix gives the formal definitions behind the Structure Score, from integrity checks and level balancing to folder economy and aggregation, and discusses refinements of reference groups and two measures of complete-group recovery.

\subsection{Integrity and Relations}

Preservation requires the original record inventory exactly once, unchanged bodies and an allowed hierarchy. An output that fails preservation receives a Structure Score of 0 and remains in its family mean. Records left in the inbox are scored as members of an inbox folder. Structural agreement is invariant under generated folder names and sibling order. For a tree T, each record i has a folder path. The shared-path depth is

\begin{equation}
d_T(i,j) = \max\left\{k \ge 0 : p_T(i)_{1:k} = p_T(j)_{1:k}\right\}
\label{eq:shared-path-depth}
\end{equation}

For every record triple, the reference branching level and the relation induced by a tree are

\begin{equation}
\begin{aligned}\ell_R(q) &= \min_{\pi\in\binom{q}{2}} d_R(\pi), \\\rho_T(q) &=\begin{cases}\displaystyle \arg\max_{\pi\in\binom{q}{2}} d_T(\pi), & \text{unique maximum}, \\\bot, & \text{otherwise}.\end{cases}\end{aligned}
\label{eq:resolved-relation}
\end{equation}

The pair variable ranges over the three record pairs in the triple.

\subsection{Level Balance}

At each reference level, count reference-resolved triples, output-resolved triples assigned to that level and matching relations:

\begin{equation}
\begin{aligned}\mathcal{Q}_\ell &= \{q : \ell_R(q)=\ell\}, \\g_\ell &= \left|\{q\in\mathcal{Q}_\ell : \rho_R(q)\ne\bot\}\right|, \\o_\ell &= \left|\{q\in\mathcal{Q}_\ell : \rho_O(q)\ne\bot\}\right|, \\m_\ell &= \left|\{q\in\mathcal{Q}_\ell : \rho_R(q)\ne\bot, \right.\\&\qquad\left.\rho_O(q)=\rho_R(q)\}\right|.\end{aligned}
\label{eq:level-counts}
\end{equation}

Recall uses every informative reference level. Precision uses the subset on which the output resolves at least one triple:

\begin{equation}
\begin{aligned}\mathcal{L}_{R} &= \{\ell : g_\ell>0\}, & R_\ell &= \frac{m_\ell}{g_\ell}, \\\mathcal{L}_{P} &= \{\ell : g_\ell>0,\ o_\ell>0\}, & P_\ell &= \frac{m_\ell}{o_\ell}.\end{aligned}
\label{eq:level-precision-recall}
\end{equation}

The metric macro-averages those level scores before taking their harmonic mean:

\begin{equation}
\begin{aligned}\bar P &= \frac{1}{|\mathcal{L}_{P}|}\sum_{\ell\in\mathcal{L}_{P}} P_\ell, \\\bar R &= \frac{1}{|\mathcal{L}_{R}|}\sum_{\ell\in\mathcal{L}_{R}} R_\ell, \\F_1^{\mathrm{bal}}(O,R) &= \frac{2\bar P\bar R}{\bar P+\bar R}.\end{aligned}
\label{eq:level-balanced-f1}
\end{equation}

If the reference resolves no triples, F1-bal and Structure Score are undefined. No benchmark task has this case. Every reference resolves triples at level 0, 148 of the 150 also resolve triples at level 1, and every task has at least 36 records. If informative levels exist but the output resolves none, F1-bal is zero. Levels lacking resolved reference triples are excluded from the macro-average. Output resolutions at those levels are reported as unsupported refinement.

\subsection{Refinement Beyond the Reference}
\label{sec:appendix.counterexample}

An output can obtain a Structure Score of 100 while splitting a reference group. Consider seven records in a two-level reference, with $a$, $b$ and $c$ in folder $A/X$, $d$ and $e$ in $B/Y$, and $f$ and $g$ in $B/Z$. The output replaces $A/X$ with two folders, $A/X1$ containing $a$ and $b$ and $A/X2$ containing $c$, and leaves $B$ unchanged. It reproduces every relation that the reference resolves: $(g_\ell,o_\ell,m_\ell)$ is $(30,30,30)$ at the parent level and $(4,4,4)$ at the leaf level. The output also resolves the triple $\{a,b,c\}$, but this triple lies at a level that the reference does not resolve and is therefore excluded. Hence $\bar P=\bar R=1$. Both hierarchies contain one single-child folder, $A$ in the reference and $X2$ in the output, so $E=1$ and the Structure Score is 100. The clade measures in Appendix~\ref{sec:appendix.clades} detect the split.
The example uses a parent with a single leaf group, a shape that occurs in 11 Thread references and in no Recap or Design reference. Nesting part of a leaf group one level deeper also leaves the Structure Score unchanged, in any family, because the added distinction lies below the leaf level of the reference.

\subsection{Clade Agreement and Exact Recovery}
\label{sec:appendix.clades}

As an additional diagnostic, we compare complete groups instead of relations. For a valid hierarchy $T$ over a record set $V$, let $L_T(v)$ be the set of records below folder $v$. The informative clades of $T$ are
\begin{equation}
\mathcal{C}_T=\{L_T(v):1<|L_T(v)|<|V|\},
\label{eq:informative-clades}
\end{equation}
that is, the record sets of all folders other than single records and the full collection; folders along a single-child chain contribute one set. Folder names and sibling order do not affect $\mathcal{C}_T$. With $k=|\mathcal{C}_O\cap\mathcal{C}_R|$, clade F1 is
\begin{equation}
F_{1,C}(O,R)=\frac{2k}{|\mathcal{C}_O|+|\mathcal{C}_R|}.
\label{eq:clade-f1}
\end{equation}
An output recovers the reference exactly when $\mathcal{C}_O=\mathcal{C}_R$. In the example of Appendix~\ref{sec:appendix.counterexample}, the output adds the clade $\{a,b\}$, which gives a clade F1 of $8/9$ and no exact recovery.

We computed both measures for a subset of runs, the ReAct-FC outputs of Claude Haiku~4.5 and GPT-6.1 Sol on Recap and Thread (Table~\ref{tab:clades}). Clade F1 is well below the Structure Score in every row. GPT-6.1 Sol reaches 35.1 on Recap and 36.1 on Thread and recovers one and two hierarchies exactly, whereas Claude Haiku~4.5 reaches 1.7 and 8.6 and recovers none.
On Design, where clade F1 was not computed, GPT-6.1 Sol recovers 30 of the 50 hierarchies exactly and Claude Haiku~4.5 recovers one.
The same Thread outputs differ in how they form parent groups. Claude Haiku~4.5 creates 11.32 top-level folders per task on average, compared with 2.62 parent issues in the reference, whereas GPT-6.1 Sol creates 2.86 but still splits 1.18 reference parent groups and mixes records from different parents in 1.10 of its top-level folders. Agreement on relations therefore does not imply that complete groups are recovered.

\begin{table}[t]
\centering
\small
\setlength{\tabcolsep}{4pt}
\begin{tabular}{@{}llrrr@{}}
\toprule
\textbf{Family} & \textbf{Model} & \textbf{\boldmath $S$} & \textbf{Clade F1} & \textbf{Exact} \\
\midrule
Recap & Claude Haiku 4.5 & 40.8 & 1.7 & 0 \\
Recap & GPT-6.1 Sol & 74.1 & 35.1 & 1 \\
Thread & Claude Haiku 4.5 & 46.2 & 8.6 & 0 \\
Thread & GPT-6.1 Sol & 77.9 & 36.1 & 2 \\
\bottomrule
\end{tabular}
\caption{Structure Score ($S$), clade F1 and exact recovery for the ReAct-FC outputs of two models on 50 tasks per family. Exact is the number of outputs that reproduce the reference hierarchy. Scores use the 0 to 100 scale.}
\label{tab:clades}
\end{table}

\subsection{Folder Economy}

The folder-economy equation uses u for the count of one-child folders and n for the count of output folders. Only one-child structure in excess of the reference count activates the correction. A wrapper can preserve all closest-pair relations while adding a navigation step; relation agreement and this folder correction therefore measure different properties. Structure Score multiplies level-balanced agreement by folder economy.

An entirely flat tree asserts zero resolved distinctions.
Filing and structure are therefore distinct: with Deep Agents, three MAI-Code-1.1-Flash Recap outputs file every record into one common folder and receive a Structure Score of 0.
 A near-correct hierarchy can retain many record relationships while differing at one group. Extra refinement, merged groups, missing records and unreadable inputs receive separate treatment. The illustrative structures describe scoring behavior; human navigation preference and downstream utility are separate constructs.

\subsection{Aggregation}

Each task-family mean assigns equal weight to its scored task instances. For each metric, Avg is the arithmetic mean of the three family means, giving Recap, Thread and Design equal weight. Structure Score is computed per task before averaging; multiplying averaged components would change the estimand. An average requires all three family values. It summarizes the contributing configurations and preserves their access, stopping and sampling qualifications.

Repeated episodes are averaged within a collection before aggregation across collections, with task-family rows retained. The exploratory design packets contain repeated proposals: 600 proposal slots draw from 294 unique proposals. Independent-workspace uncertainty therefore uses source collections instead of triples or record appearances.


\section{Task-Level Results}
\label{app:tables}
\label{app:pertask}

\begin{figure*}[!t]
\centering
\includegraphics[width=\textwidth]{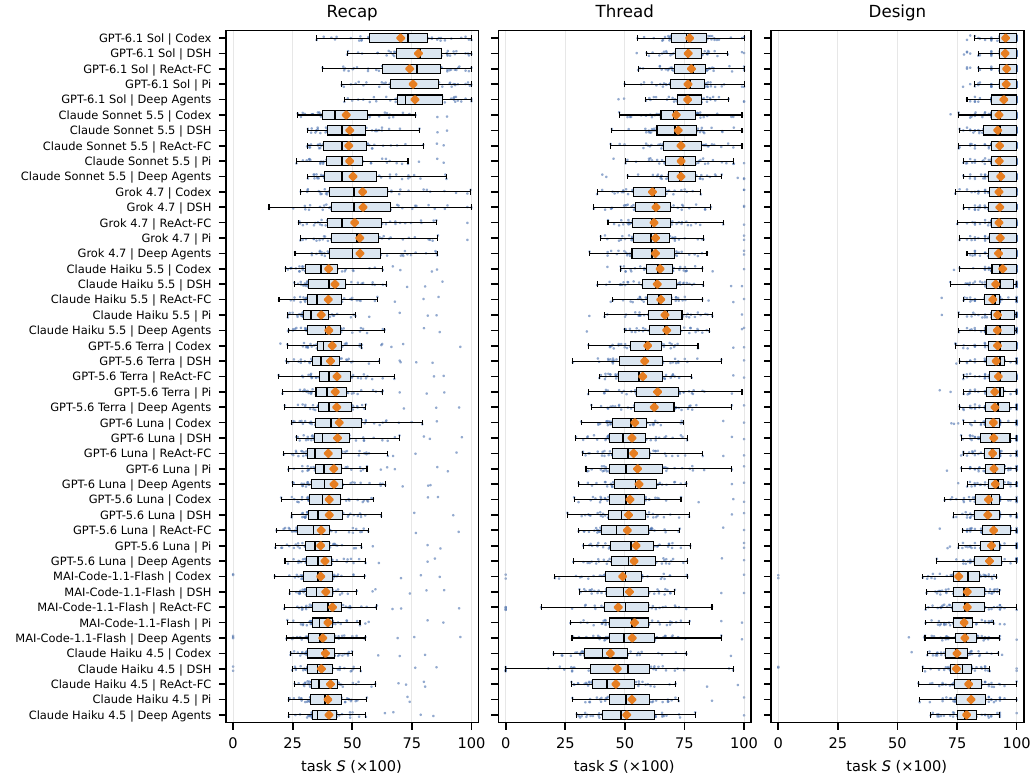}
\caption{Per-task structure score of every complete configuration (rows ordered by model mean, then harness). Boxes show the interquartile range and median, orange diamonds the mean, dots individual tasks.}
\label{fig:pertask}
\end{figure*}

\begin{figure*}[!t]
\centering
\includegraphics[width=\textwidth]{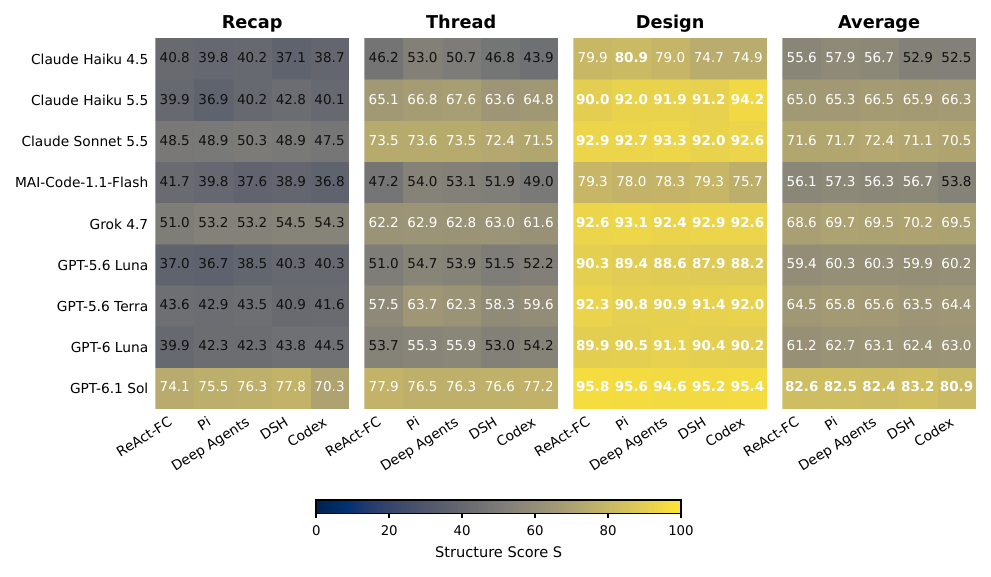}
\caption{Structure Score of the 45 configurations in Table~\ref{tab:T5} by model, harness and task family. Read each row to compare harnesses for one model and each column to compare models under one harness. The Average panel reports the unweighted mean of the three family scores.}
\label{fig:appendix-score-heatmaps}
\end{figure*}

\begin{figure*}[!t]
\centering
\includegraphics[width=0.9\textwidth]{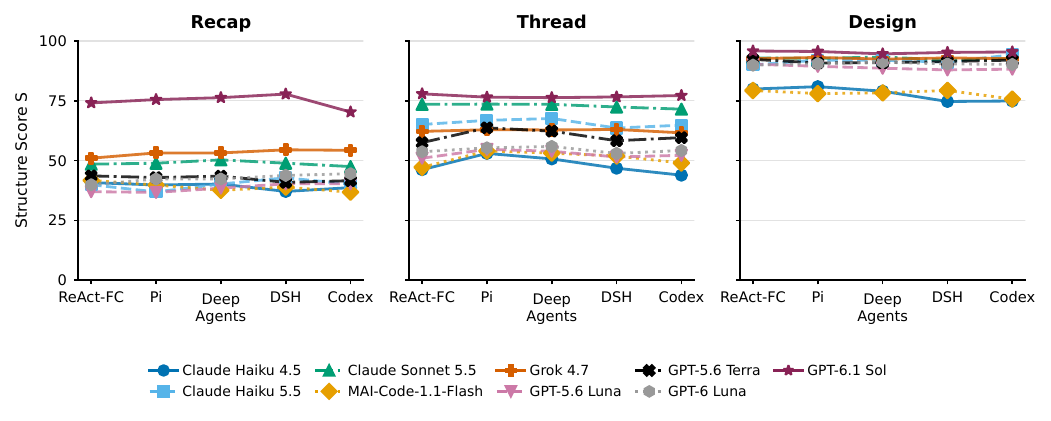}
\caption{Structure Score of the 45 configurations in Table~\ref{tab:T5} as model profiles across harnesses. Each line fixes the model and varies the harness; each vertical slice compares models under the same harness.}
\label{fig:appendix-score-profiles}
\end{figure*}

Table~\ref{tab:model-means} summarizes each model's mean $S$ across its configurations in Table~\ref{tab:T5}. Figure~\ref{fig:pertask} gives the per-task scores behind each configuration mean. Figures~\ref{fig:appendix-score-heatmaps} and~\ref{fig:appendix-score-profiles} arrange the same scores by model and harness.
Recap and Thread scores are spread across most of the scale within every configuration, while Design scores cluster near the top with a short tail of low tasks.

\noindent\begin{minipage}{\linewidth}
\captionsetup{type=table}
\centering
\small
\begin{tabular}{@{}lr@{}}
\toprule
\textbf{Model} & \textbf{\boldmath Mean $S$} \\
\midrule
GPT-6.1 Sol        & \textbf{82.3} \\
Claude Sonnet 5.5  & 71.5 \\
Grok 4.7           & 69.5 \\
Claude Haiku 5.5   & 65.8 \\
GPT-5.6 Terra      & 64.7 \\
GPT-6 Luna         & 62.5 \\
GPT-5.6 Luna       & 60.0 \\
MAI-Code-1.1-Flash & 56.0 \\
Claude Haiku 4.5   & 55.1 \\
\bottomrule
\end{tabular}
\caption{Mean $S$ ($\times 100$) of each model over its configurations in Table~\ref{tab:T5}.}
\label{tab:model-means}
\end{minipage}\par\medskip

\section{Text-Only Baselines}
\label{app:baselines}

The baselines use only the record text of each packet and have no access to the reference. Each baseline represents records by within-packet TF-IDF vectors over word unigrams and bigrams \citep{ref-irbook,ref-scikit} and builds one average-linkage dendrogram over cosine distances. Fixed cuts take the parent and leaf groups at distances 0.9 and 0.7. Silhouette cuts choose the numbers of parent and leaf groups that maximize the mean silhouette of the two cuts \citep{ref-silhouette}, with at least twice as many leaf groups as parent groups. Random shape keeps the silhouette hierarchy but assigns records to its positions at random, with five seeds per task. Filed flat places all records in one folder. By construction, all 1,200 baseline outputs keep and file every record. Table~\ref{tab:baselines} reports their Structure Scores.

\begin{table}[t]
\centering
\small
\begin{tabular}{@{}lrrr@{}}
\toprule
\textbf{Baseline} & \textbf{Recap} & \textbf{Thread} & \textbf{Design} \\
\midrule
Filed flat & 0.0 & 0.0 & 0.0 \\
Fixed cuts & 9.4 & 16.6 & 45.9 \\
Random shape & 18.5 & 19.6 & 17.0 \\
Silhouette & 26.9 & 34.0 & 71.0 \\
\bottomrule
\end{tabular}
\caption{Structure Score of the text-only baselines on 50 tasks per family. Random shape averages five seeds per task.}
\label{tab:baselines}
\end{table}

Silhouette cuts score higher than fixed cuts on all 50 tasks of each family, and higher than the mean of the random shapes on 48 Recap tasks and on all Thread and Design tasks. They select the largest permitted number of parent groups in 36 Recap, 35 Thread and 47 Design tasks, producing 7.12, 7.06 and 5.94 parent groups on average, compared with 2.68, 2.62 and 4.00 in the reference. The baseline therefore over-segments, and its scores serve as reference points rather than estimates of attainable performance. No baseline output recovers a reference hierarchy exactly. Random shapes score above zero because random groupings still resolve some reference triples correctly, so a positive Structure Score alone does not indicate a meaningful organization. With $\lambda$ set to 0 and 0.4, the silhouette scores are 28.7 and 25.1 on Recap, 37.2 and 30.8 on Thread, and 73.9 and 68.1 on Design.
Removing issue keys, pull-request numbers, KEP numbers and SIG names from the Thread text before clustering lowers the silhouette score from 34.0 to 32.7, so these identifiers contribute little to the lexical baseline.

\section{Coverage of Configurations}
\label{app:coverage}

Table~\ref{tab:T5} reports all 45 pairings of the nine models and five harnesses, and every family cell covers all 50 tasks of the family. Gemini~4 Argon is not included because its runs do not cover all tasks.
Intervals are computed by resampling tasks.

\section{Error Definitions and a Worked Example}
\label{app:defs}

\paragraph{Triples and levels.}
The scorer compares the reference and output trees on triples of records. A triple is \emph{reference-distinguishable} when two of its records share a deeper folder in the reference than either shares with the third. Its \emph{level} is set by the reference: \emph{parent} if the three records are not all under the same parent (e.g.\ two in one SIG, one in another), \emph{leaf} if they share a parent but not a leaf (e.g.\ two in one KEP, one in a sibling KEP).

\paragraph{Outcomes.}
For each triple of records that the reference distinguishes, the output is \emph{agreed} if it pairs the same two records more closely, \emph{conflicting} if it pairs a different two, and \emph{unresolved} if it places all three at the same depth. Conflicts express a wrong relation and lower precision and recall; unresolved triples express a missing relation and lower recall.

\paragraph{Score.}
$S=F_1^{\mathrm{bal}}\times E$ (Equation~\ref{eq:structure-score}), where $F_1^{\mathrm{bal}}$ averages triple precision and recall across levels before combining them, so that the many parent-level relations do not outweigh the fewer leaf-level ones, and the economy factor
\[E = 1 - 0.2\times\frac{\text{excess single-child folders}}{\text{output folders}}\]
penalizes empty shells. Folder names, folder order and a wrapper folder shared by all records do not affect $S$. Economy accounts for 1--10\% of the shortfall $1-S$ depending on the family; the rest is relation recovery.

\paragraph{Output categories.}
Each output receives one exclusive category: \emph{none} (no structural errors; same tree up to names and order), \emph{parent} (all leaf-level triples agreed, some parent-level triple not), \emph{leaf} (the reverse), or \emph{both}. Table~\ref{tab:levels} reports the triple outcomes and categories by family.

\begin{figure*}[!t]
\centering
\includegraphics[width=0.92\textwidth]{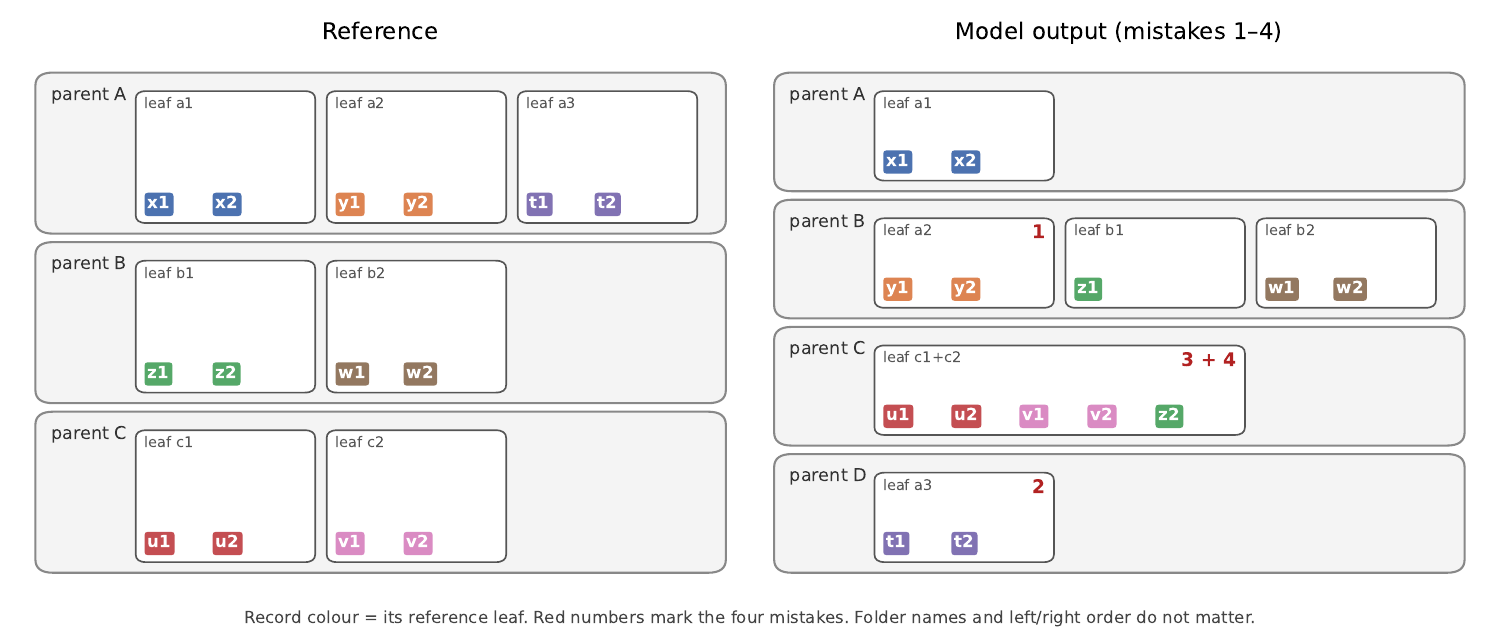}
\caption{One reference and one output containing every error case. Record colour marks its reference leaf; red numbers mark the four mistakes discussed in Appendix~\ref{app:defs}.}
\label{fig:toy}
\end{figure*}

\paragraph{Worked example.}
Figure~\ref{fig:toy} shows one reference and one output that contains every case at once.
Moving leaf a2 under B (mistake~1) makes $\{x_1,y_1,z_1\}$ pair $y_1$--$z_1$ instead of $x_1$--$y_1$: a conflicting parent-level triple.
Putting leaf a3 alone in a new parent D (mistake~2) leaves $\{x_1,t_1,z_1\}$ in three different parents: an unresolved parent-level triple, the pattern of a KEP isolated from its SIG.
Merging c1 and c2 (mistake~3) leaves $\{u_1,u_2,v_1\}$ in one leaf: an unresolved leaf-level triple.
Moving the single record $z_2$ (mistake~4) breaks both levels at once.
Over all triples, the output has 104 agreed, 42 conflicting and 94 unresolved parent-level triples and 14 / 2 / 4 leaf-level ones ($S=0.58$); counts were verified against the released scorer.

\begin{table*}[!t]
\centering
\small
\setlength{\tabcolsep}{4pt}
\begin{tabular}{@{}llrrrrrrrrrrr@{}}
\toprule
& & & \multicolumn{3}{c}{\textbf{Parent-level triples}} & \multicolumn{3}{c}{\textbf{Leaf-level triples}} & \multicolumn{4}{c}{\textbf{Outputs by category}} \\
\cmidrule(lr){4-6}\cmidrule(lr){7-9}\cmidrule(lr){10-13}
\textbf{Family} & \textbf{Parent / leaf} & \textbf{Configs} & \textbf{agr.} & \textbf{conf.} & \textbf{unres.} & \textbf{agr.} & \textbf{conf.} & \textbf{unres.} & \textbf{none} & \textbf{parent} & \textbf{leaf} & \textbf{both} \\
\midrule
Recap & meeting / section & 25 & 0.38 & 0.35 & 0.27 & 0.46 & 0.36 & 0.18 & 3 & 0 & 33 & 1214 \\
Thread & parent issue / subtask & 21 & 0.40 & 0.07 & 0.54 & 0.68 & 0.09 & 0.22 & 16 & 6 & 12 & 1016 \\
Design & SIG / KEP & 25 & 0.87 & 0.02 & 0.12 & 0.99 & 0.00 & 0.01 & 304 & 596 & 52 & 298 \\
\bottomrule
\end{tabular}
\caption{Triple outcomes by level (median share over the configurations with per-level data) and the exclusive category of each output, counted over configurations with per-level data for all 50 tasks (Recap 25, Thread 21, Design 25). \emph{None}: no structural errors (identical tree up to folder names and order); \emph{parent}: every leaf-level triple agreed and some parent-level triple did not; \emph{leaf}: the reverse; \emph{both}: disagreements at both levels (including 5 Thread outputs that agree on all triples but differ in tree shape). Definitions in Appendix~\ref{app:defs}.}
\label{tab:levels}
\end{table*}

\paragraph{GPT-6.1 Sol versus other configurations.}
\label{app:sol}
Two Sol configurations (Codex and ReAct-FC) have per-level data for all tasks. Their parent-level agreement is 0.74--0.79 on Recap, 0.74--0.76 on Thread and 0.93--0.94 on Design, against medians of 0.38, 0.38 and 0.86 over the other configurations; their unresolved parent-level share on Thread is 0.18--0.20 against 0.54. Leaf-level agreement rises less: 0.68--0.71 against 0.46 on Recap and 0.80--0.81 against 0.68 on Thread. Each parent-level comparison has raw $p<0.01$ (Mann--Whitney over 2 versus 19--23 configurations) but not after Holm correction.

\section{Inspected Examples}
\label{app:examples}

All three examples come from GPT-5.6 Terra with ReAct-FC and were chosen by a fixed rule (the task at the 10th percentile of $S$ within the family), not by inspection.

\paragraph{Recap, grouping by the wrong dimension (mini-050, $S=0.33$).}
The reference has 4 meetings with 15 summary sections. The output builds 3 folders named after project phases (kickoff, concept review, final design) with 9 leaves; 8 of the 9 leaves mix notes from 2--4 meetings, none of the 15 reference sections is recovered, and the largest leaf holds 20 notes. The phases are coherent topics, but they cut across meetings.

\paragraph{Thread, missing parent layer (mini-026, $S=0.43$).}
The reference has 2 parent issues with 13 subtasks. The output creates 13 top-level folders, one per subtask, each with a single child and no parent layer, and 3 of the 13 subtasks are recovered. The same model with Codex builds the two parent folders for this task.

\paragraph{Design, parent-level error (mini-044, $S=0.83$).}
The reference has 4 SIGs and 12 KEPs of three excerpts each. The output reproduces all 12 KEPs and attaches \emph{immutable secrets and config maps} (storage) to apps, producing 162 conflicting parent-level triples, and places \emph{DRA partitionable devices} (scheduling) in a new node folder of its own, producing unresolved triples (891 in total with the first error). The leaf level is fully correct.


\section{Controlled Variations}
\label{app:variations}

Table~\ref{tab:variations} reports three identical medium-effort runs and one high-effort run of GPT-5.6 Terra with Codex. Per-task changes follow a few discrete structural decisions: on Thread, 76\% of large per-task changes (more than 10 points) coincide with building or not building a parent layer. Its per-task changes relative to the medium runs are no larger than those between identical medium runs.

\section{Task Features and Difficulty}
\label{app:features}

Table~\ref{tab:features} correlates task features with each task's mean score over the 45 configurations. On Thread, structural features remain significant after correction: tasks with more subtasks, smaller subtasks or more low-information comments score lower, and lexically separable subtasks score higher. Recap scores rise with how lexically distinct the meetings are. On Design, no feature is significant, and the lowest- and highest-scoring tasks match on record count, structure, length and lexical separability.

\begin{table}[!t]
\centering
\small
\setlength{\tabcolsep}{4pt}
\begin{tabular}{@{}lrrr@{}}
\toprule
\textbf{Task feature} & \textbf{Recap} & \textbf{Thread} & \textbf{Design} \\
\midrule
Records & $-$0.34$^{*}$ & +0.40$^{*}$ & n/a \\
Parents & $-$0.24 & n/a & n/a \\
Leaves & $-$0.27 & $-$0.60$^{**}$ & n/a \\
Mean leaf size & +0.13 & +0.69$^{**}$ & n/a \\
Mean record length & $-$0.13 & +0.23 & $-$0.18 \\
Short-record share & +0.08 & $-$0.40$^{*}$ & n/a \\
Low-information share & $-$0.31$^{*}$ & $-$0.49$^{**}$ & n/a \\
Parent separability & +0.48$^{**}$ & +0.29$^{*}$ & +0.04 \\
Leaf separability & +0.33$^{*}$ & +0.67$^{**}$ & $-$0.25 \\
\bottomrule
\end{tabular}
\caption{Spearman $\rho$ between a task feature and the task's mean $S$ over the 45 configurations (50 tasks per family). Parents/leaves: counts in the reference; separability: lexical (TF-IDF) distinctness of reference groups; n/a: not tested (constant or nearly constant within the family). $^{**}$Holm-corrected $p<0.05$ across all 31 feature tests; $^{*}$raw $p<0.05$.}
\label{tab:features}
\end{table}

\paragraph{Identical Records.}
\label{app:ambiguity}
Five Recap tasks and 22 Thread tasks contain records with identical text but different reference paths; Design tasks contain none. For each triple resolved by the reference, we exchange the reference paths of two such records, one pair at a time, and mark the triple as changed if any exchange alters its closest pair or leaves it unresolved. All other record fields and group sizes stay fixed. Table~\ref{tab:ambiguity} reports the results for the tasks in Table~\ref{tab:T2}. In Thread, changes occur only at the leaf level: identical records with different paths always belong to the same parent issue, so none of the 70,513 parent-level triples that involve such a record changes. Because similar but non-identical records are not exchanged, these shares are a lower bound on ambiguity from record content. The check does not change any score.

\begin{table}[t]
\centering
\footnotesize
\setlength{\tabcolsep}{2pt}
\begin{tabular*}{\columnwidth}{@{\extracolsep{\fill}}llrrrrr@{}}
\toprule
\textbf{Family} & \textbf{Level} & \textbf{Tasks} & \textbf{Triples} & \textbf{Changed} & \textbf{Pooled} & \textbf{Mean} \\
\midrule
Recap & Parent & 50 & 639,186 & 2,178 & 0.34 & 0.29 \\
Recap & Leaf & 50 & 93,275 & 1,690 & 1.81 & 1.56 \\
Thread & Parent & 50 & 591,197 & 0 & 0.00 & 0.00 \\
Thread & Leaf & 48 & 110,066 & 6,511 & 5.92 & 7.67 \\
Design & Parent & 50 & 194,400 & 0 & 0.00 & 0.00 \\
Design & Leaf & 50 & 10,800 & 0 & 0.00 & 0.00 \\
\bottomrule
\end{tabular*}
\caption{Reference triples whose relation changes when the reference paths of two records with identical text are exchanged. Tasks is the number of tasks with resolved reference triples at that level, and Triples is the number of those triples. Pooled is the percentage of all triples that change, and Mean is the average of the per-task percentages.}
\label{tab:ambiguity}
\end{table}

\paragraph{Record Screen and Visible Identifiers.}
The record-quality screen in Section~\ref{sec:task-properties} lowercases each record and collapses its whitespace, then assigns the first matching flag: identical text under different reference paths, release bookkeeping (for example, ``Bulk update: moved all 3.2.0 non-blocker issues''), or acknowledgement only (for example, ``+1. committed to branch-2''). In Thread, it flags 108 duplicate, 50 bookkeeping and 56 acknowledgement appearances in 22, 28 and 30 tasks, and 42 tasks have at least one flag. In Recap, it flags 16 duplicate appearances in five tasks. A separate scan counts explicit identifiers that could reveal the reference grouping: Jira issue keys, pull-request numbers or links, KEP numbers and SIG names. Thread records contain 561 issue keys and 618 pull-request references, Design records contain one pull-request reference and no KEP numbers or SIG names, and Recap records contain none. The scan does not detect less explicit cues.

\section{Quality and Token Use}
\label{app:cost}

Table~\ref{tab:cost} pairs each model's quality with its token use per task. GPT-6.1 Sol's higher Recap score comes with the highest token use, while on Thread Claude Sonnet~5.5 comes within about three points of Sol with a fifth of its tokens; MAI-Code-1.1-Flash uses many tokens for low scores.
Appendix~\ref{sec:appendix.cost} reports run-level provider calls and token totals.

\begin{table}[!t]
\centering
\small
\setlength{\tabcolsep}{3pt}
\begin{tabular}{@{}lrrrrrr@{}}
\toprule
& \multicolumn{2}{c}{\textbf{Recap}} & \multicolumn{2}{c}{\textbf{Thread}} & \multicolumn{2}{c}{\textbf{Design}} \\
\cmidrule(lr){2-3}\cmidrule(lr){4-5}\cmidrule(lr){6-7}
\textbf{Model} & \textbf{\boldmath $S$} & \textbf{kTok} & \textbf{\boldmath $S$} & \textbf{kTok} & \textbf{\boldmath $S$} & \textbf{kTok} \\
\midrule
GPT-6.1 Sol & 76.5 & 305 & 76.4 & 979 & 95.1 & 342 \\
Claude Sonnet 5.5 & 49.4 & 72 & 73.2 & 185 & 92.7 & 86 \\
Grok 4.7 & 53.8 & 184 & 62.5 & 490 & 92.7 & 94 \\
Claude Haiku 5.5 & 40.0 & 85 & 65.6 & 247 & 91.9 & 84 \\
GPT-5.6 Terra & 42.7 & 47 & 60.8 & 146 & 91.5 & 51 \\
GPT-6 Luna & 42.6 & 57 & 54.4 & 187 & 90.4 & 51 \\
GPT-5.6 Luna & 38.5 & 45 & 53.4 & 160 & 88.6 & 48 \\
MAI-Code-1.1-Flash & 38.8 & 363 & 53.0 & 848 & 78.7 & 260 \\
Claude Haiku 4.5 & 39.0 & 147 & 50.2 & 402 & 78.6 & 95 \\
\bottomrule
\end{tabular}
\caption{Quality and token use per task, over configurations whose usage is recorded for all 50 tasks of the family: mean $S$ ($\times100$) and the median over configurations of mean tokens per task (thousands). Token accounting differs across harnesses (e.g.\ cached input), so cross-harness token gaps are indicative.}
\label{tab:cost}
\end{table}

\end{document}